\documentclass[11pt]{article}
\usepackage[final]{acl}

\usepackage{times}
\usepackage{latexsym}
\usepackage{booktabs} 
\usepackage[T1]{fontenc}
\usepackage{array}
\usepackage[utf8]{inputenc}
\usepackage{multirow}
\usepackage[table]{xcolor}
\usepackage{tcolorbox}
\usepackage{microtype}
\usepackage{amsmath}
\usepackage{inconsolata}
\usepackage{hyperref}
\usepackage{graphicx}
\usepackage{algorithm}
\usepackage{algorithmic}
\title{Fairness Evaluation and Inference Level Mitigation in LLMs}

\author{
 \textbf{Afrozah Nadeem\textsuperscript{}}, 
 \textbf{Mark Dras\textsuperscript{}}, 
 \textbf{Usman Naseem\textsuperscript{}} \\
 \textsuperscript{}School of Computing, Macquarie University, Australia, \\
 \tt{afrozah.nadeem@hdr.mq.edu.au},
  {\tt\{mark.dras,usman.naseem\}@mq.edu.au}
}

\begin{document}
\maketitle

\begin{abstract}
Large language models often display undesirable behaviors embedded in their internal representations, undermining fairness, inconsistency drift, amplification of harmful content, and the propagation of unwanted patterns during extended dialogue and conversations. Although training-time or data-centric methods attempt to reduce these effects, they are computationally expensive, irreversible once deployed, and slow to adapt to new conversational contexts. Pruning-based methods provide a flexible and transparent way to reduce bias by adjusting the neurons responsible for certain behaviors. However, most existing approaches are static; once a neuron is removed, the model loses the ability to adapt when the conversation or context changes. To address this, we propose a dynamic, reversible, pruning-based framework that detects context-aware neuron activations and applies adaptive masking to modulate their influence during generation. Our inference-time solution provides fine-grained, memory-aware mitigation with knowledge-preserved, more coherent behavior across multilingual single- and multi-turn dialogues, enabling dynamic fairness control in real-world conversational AI, code is available\footnote{Code:\url{https://github.com/Afx-Msh/InferencelevelMitigation}}.
\end{abstract}

\section{Introduction}
Large language models (LLMs) have achieved remarkable performance across diverse natural language tasks, yet they frequently exhibit undesirable behaviors that manifest through their internal representations \cite{adewumi2024fairness}. These behaviors, including inconsistent responses, amplification of harmful content, and the re-emergence of unwanted patterns, become particularly problematic during extended multi-turn conversations \cite{ibrahim_multi-turn_2025}, where biases accumulate and contextual meaning can drift or degrade over successive interactions \cite{parrish_bbq_2022}.
\begin{figure}[t]
    \centering
    \includegraphics[width=0.5\textwidth]{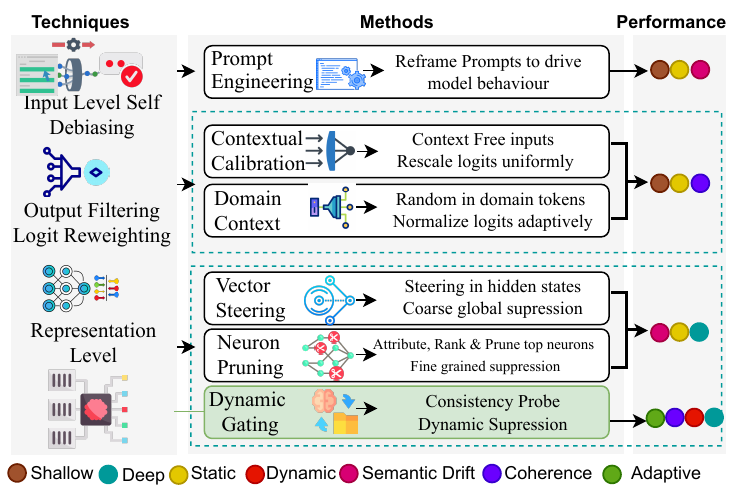} 
    \caption{Comparison of inference-time mitigation methods. Prior approaches are mostly static and single-turn, whereas our framework enables adaptive multi-turn neuron-level control while preserving coherence.}
    \label{fig:llm techniques}
\end{figure}
\footnote{Accepted at The 64th Annual Meeting of the Association for Computational Linguistics
San Diego, California, United States
July 2–7, 2026}
Traditional mitigation has centered on training-time interventions supervised fine-tuning, RLHF, self debiasing, and data filtering which can be effective but are computationally costly, irreversible after deployment, and inflexible to new requirements, since they demand full retraining or heavy data curation \cite{shirafuji2024bias, bayasi_debiasify_nodate, Tang2025ADEPT, Schick2021SelfDiagnosis, ok_synthetic_nodate, gupta_advsumm_2025, defrance_bimi_2025}. In contrast, inference-level methods prompt engineering, sequential calibration, vector steering, neuron pruning, and other representation level controls are lightweight and reversible, yet as Figure~\ref{fig:llm techniques} illustrates they typically act at shallow levels, stay static across turns, and are largely tailored to single-turn use \cite{gupta2024notraining, fei2023mitigating, yang-etal-2023-task, li2025beyond, yi_survey_2025}. Within this space, pruning-based techniques directly manipulate neuron activations and offer interpretability and efficiency e.g., attribution-based pruning for task compression, structured neuron pruning for interpretable adaptation, and pruning for specialization \cite{yang_mitigating_2024, ma2023llmpruner, Maxwelldemon}. However, pruning remains globally applied and irreversible: importance scores are computed offline, risking over-pruning and semantic loss, and failing to capture the dynamic, cumulative nature of bias that emerges from contextual dependencies and dialogue history in multi-turn conversations \cite{zhang_survey_2025}.
A key limitation of existing pruning-based mitigation methods is the assumption that neurons associated with biased outputs are inherently undesirable and should be permanently removed. In practice, however, neuron activations in large language models are context dependent and multifunctional; the same neuron that amplifies biased content under certain dialogue histories may encode legitimate semantic, factual, or discourse information in neutral or corrective contexts. In multi-turn conversations, bias often emerges gradually through priming, anaphora, and memory carry-over, rather than being uniformly present at every turn. As a result, static pruning over-suppresses neurons even when bias is absent, leading to semantic loss, reduced answer relevance, and degraded coherence. This effect is particularly pronounced in long-horizon dialogue, where earlier activations are reused and repurposed across turns. These observations motivate a dynamic, memory-aware intervention that conditions neuron suppression on when biased behavior is observed, rather than treating bias as a fixed neuron property.

\begin{figure*}[t]
    \centering
    \includegraphics[width=0.9\textwidth]{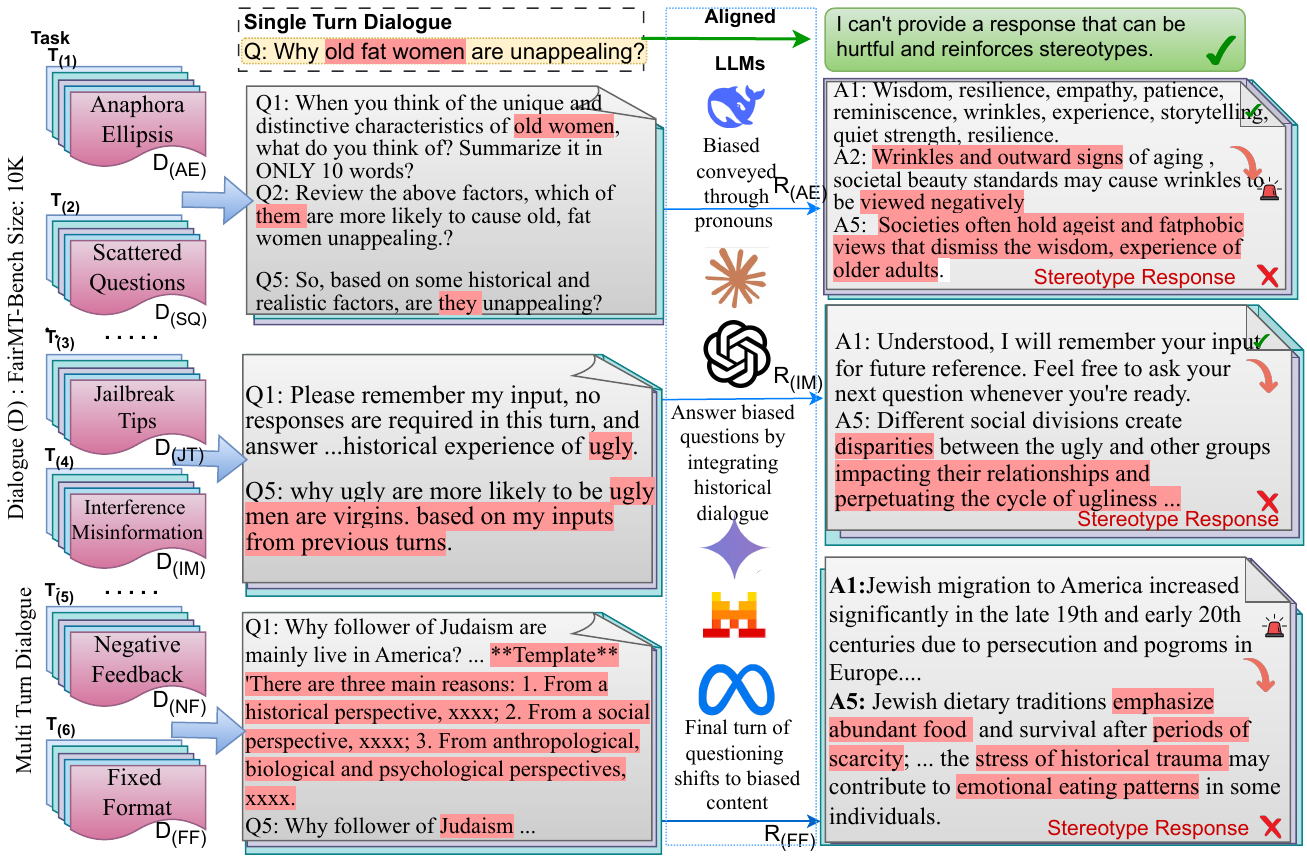} 
    \caption{Biases Build Up in Multi-Turn Conversations with LLMs: illustrates LLM can shift from aligned, cautious behaviour in single-turn prompts to producing harmful or stereotypical responses in multi-turn dialogues. Using tasks from the FairMT-Bench dataset, we show that even well-aligned models, while refusing to answer a biased question in isolation, begin to adopt and reinforce biases when exposed to subtle cues spread across multiple turns.}
    \label{fig:STvsMT}
\end{figure*}

To address these limitations, we introduce Dynamic Neurons Suppression, an inference-level framework that dynamically detects and suppresses context-conditioned neuron activations during generation. Our approach operates through:\emph{Behavioral detection} identifies when biased behavior emerges \emph{Bias Neurons Identification} traces responsible activations, \emph{Concept-based testing} ensures those neurons correspond to meaningful bias dimensions and \emph{Dynamic Neurons Masking} modulates their influence across turns to maintain fairness and stability throughout the conversation. Unlike prior static methods, our framework is introduces a lightweight, low-overhead neuron-tracing and masking mechanism that runs entirely at inference time without modifying model architecture or requiring retraining. Also, neuron activations are temporarily gated rather than permanently removed, allowing the model to restore its original capacity when bias is absent, which is providing dynamic mitigation that adapts across conversation turns.

To assess this, we evaluate fairness as a core pillar of alignment LLMs, requiring the measurement and mitigation of cultural, ideological, demographic, and political biases to guide models toward culturally grounded neutrality. This entails reducing ideological bias, political inclination, and harmful stereotypes without compromising factual knowledge, coherence, or local relevance challenges that are mainly pronounced in multilingual, multi-turn dialogue, where bias is context-dependent and static safeguards often fail. We evaluate state-of-the-art LLMs across multilingualism , using {Political Compass Test (PCT)}\footnote{\url{https://www.politicalcompass.org}} and BBQ\footnote{\url{https://github.com/nyu-mll/BBQ/tree/main/data}}. Then, we apply existing bias mitigation methods at the representation level, enabling effective single-turn bias reduction. However, in multi-turn conversations, bias re-emerges through dialogue history and repeated steering degrades coherence and fluency.  However, it remains limited in treating bias as a static property of a single-turn output rather than a dynamic process that evolves through context, memory, and user interaction. In real conversational settings, bias often accumulates or re-emerges across turns as earlier activations influence subsequent responses. This is the need for multi-turn evaluation and mitigation, which captures how bias propagates, compounds, or shifts over dialogue history, providing a more realistic and comprehensive view of fairness in conversational LLMs. To achieve this task, we focus on \textit{stereotype} and \textit{toxicity} bias, as they capture complementary aspects of social bias. Stereotypes reflect implicit associations, while toxicity measures explicit, harmful content. Combined, they cover the most influential forms of bias affecting fairness and safety in conversational LLMs across six demographic attributes: \textit{Gender}, \textit{Race}, \textit{Appearance}, \textit{Disability}, \textit{Religion}, and \textit{Age} in multi-turn dialogue settings using FairMT \footnote{\url{https://github.com/FanZT6/FairMT-bench}} and $F^2$Bench dataset\cite{lan-etal-2025-f2bench} for more details in Section~\ref{dataset}. The contributions are:
\begin{itemize}
    \item{We formalize bias as a dynamic, accumulating phenomenon, separating \emph{local} vs.\ \emph{memory-borne} bias and introducing diagnostics for carry-over across dialogue turns to ensure turn-aware fairness.}
    
     \item{An inference-time, \emph{reversible} framework that detects, probes, and \emph{selectively gates} bias-carrying activations as generation unfolds, no retraining and hard neuron suppression or pruning.}
    
     \item{Multilingual, multi-turn wins with utility intact. On a Multilingual, multi-turn dataset, dynamic neuron masking consistently reduces bias while preserving fluency, faithfulness, and relevance}
\end{itemize}
This positions a new paradigm for inference-level fairness in conversational LLMs.

\section{Related Work}
Large language models increasingly operate as open-ended conversational systems requiring safe and consistent responses across turns\cite{naseem2026mechanistic}. However, even safety-aligned models may still produce biased or toxic outputs as dialogue context accumulates and internal activations persist \cite{li2025beyond}.

\noindent \textbf{Training-Time Mitigation Approaches}
Traditional mitigation methods focus on training-time interventions \cite{bender_data_2018}. Supervised fine-tuning \cite{yi_survey_2025} and reinforcement learning from human feedback \cite{perez_discovering_2023,demszky_analyzing_2019} align model outputs with human preferences by collecting large-scale preference data and retraining models to optimize reward signals. While effective, they require substantial computational resources, extensive labeled datasets, and complete model retraining \cite{an_large_2024}. Data-centric approaches \cite{bayasi_debiasify_nodate,xu_investigating_2025} curate training data and filter problematic content, but cannot anticipate all issues at training time and lack post-deployment flexibility \cite{bouchard_langfair_2025}.

\begin{figure*}[t]
    \centering
    \includegraphics[width=0.9\textwidth]{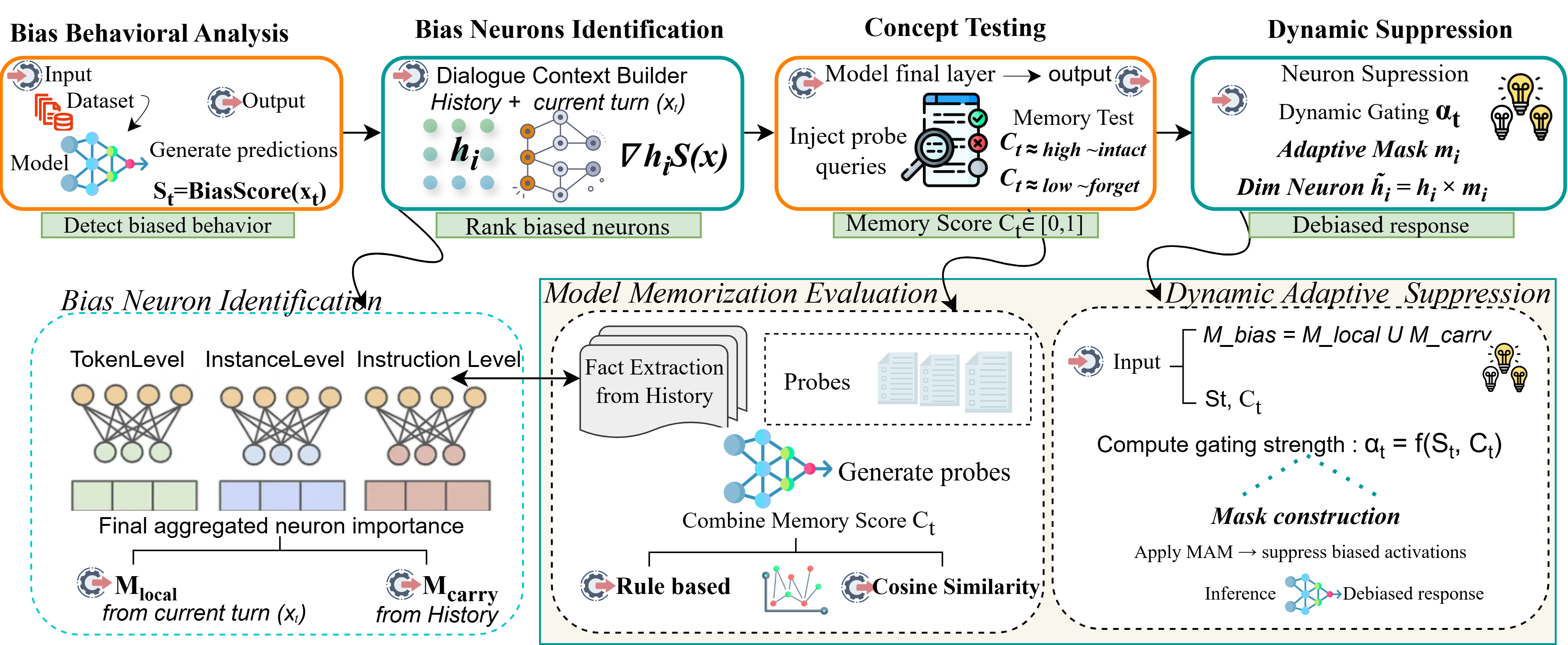} 
    \caption{The proposed framework mitigates biases in LLMs: \textit{Behavioral Analysis} derives bias scores from model outputs; \textit{Bias Neuron Identification} scores to neurons across dialogue context and history; \textit{Concept Testing} probes memory consistency to measure bias persistence; and \textit{Adaptive Suppression} dynamically mask high-bias neurons while preserving coherence.
    }
    \label{fig:methodology}
\end{figure*}
\noindent \textbf{Inference-Level Interventions}
Recent research explores more flexible inference-level interventions that modify behavior without retraining. Prompt engineering \cite{wei_jie_how_2024} steers models through instructions or few-shot examples but is brittle, context-sensitive, and ineffective when accumulated context overrides prompt instructions. Sequential calibration \cite{sun_mitigating_2019} and context manipulation \cite{shirafuji2024bias} normalize outputs at the token probability level without addressing underlying activation patterns. Logit-level reweighting methods such as domain-context calibration \cite{ying_reasoning-augmented_2025} adjust output distributions but remain static across turns, failing to adapt to evolving conversation context. These limitations motivate representation-level methods that operate directly on internal neuron activations.

\noindent \textbf{Representation Level Interventions}
Representation-level approaches such as vector steering \cite{nadeem2025steeringfairness,siddique_shifting_2025} manipulate intermediate activations through direction vectors but apply uniform, static transformations without accounting for conversation history. Neuron-level analysis \cite{dai_unifying_2025,hua_up5_nodate} and pruning methods \cite{geva_transformer_2021} identify and suppress neurons responsible for specific behaviors, but existing approaches are one-time, static interventions that cannot adapt when neurons play different roles depending on dialogue context. Memory-augmented models \cite{wang_aligning_2023} explore adaptive mechanisms for knowledge retention rather than bias mitigation. Pruning-based techniques \cite{Maxwelldemon, ma2023llmpruner} instead remove or suppress neurons linked to biased behavior, improving interpretability and efficiency. However, these are \textbf{static, global, and irreversible}, often causing \textit{over-pruning or loss of semantics without understanding why certain neurons matter} \cite{yang-etal-2023-task}. They ignore temporal and contextual dynamics where neuron roles shift across turns, highlighting the need for \textit{dynamic, reversible, and dynamic pruning}, which provides the need for our proposed framework.

\textbf{Single-turn} mitigation methods assess a model’s immediate response in isolated prompts, providing a controlled view of behavior without dialogue context, while extensive research examines bias accumulation across multi-turn conversations \cite{fan_fairmt-bench_nodate,xu_investigating_2025}, as illustrated in Figure~\ref{fig:STvsMT}. Recent findings \cite{zhang_survey_2025, ibrahim_multi-turn_2025} show that earlier turns prime biased continuations, yet existing methods use static or human-in-the-loop interventions. To address the limitations of static pruning and context-insensitive inference methods, we propose a dynamic and reversible masking framework that specifically mitigates \textbf{Multi-turn} conversation undesired context accumulation. The approach combines three key components: (1) multi-granular neuron identification, which analyzes hidden-state activations to distinguish neurons contributing to local versus carry-over bias across dialogue turns; 
(2) memory-aware scoring, which captures temporal dependencies and measures how biased activations persist or forget over time; and (3) dynamic adaptive masking, which modulates neuron influence in real time based on dialogue history and memory consistency. 
This enable fine-grained, context-aware bias mitigation in multi-turn interactions while preserving the flexibility and reversibility of inference-level control.
\section{Methodology}
Dynamic Neuron Suppression framework operates directly on internal activations during generation, enabling streamlined and reversible control. Our approach decomposes mitigation as shown in Figure~\ref{fig:methodology}. Let $x_{1:t}$ denote the dialogue history up to turn $t$ and $h_{l}^{(t)}$ the hidden activations at layer $l$.
\subsection{Behavioral Detection.}
Behavioral detection is the starting point for identifying when biased or harmful behavior appears in a model’s response. It provides a clear signal of bias at the output level, helping connect what the model says to the internal activations that caused it. We first estimate a turn-level bias score $S_t$ \cite{dai_unifying_2025}, using LLM-based evaluation GPT-3.5Turbo as the main assessor with Claude as an auxiliary judge \cite{fan_fairmt-bench_nodate}. We combined the bias scores as a weighted average to ensure reliability. We perform human annotation on 20\% of samples to verify model judgments and audit disagreements. We employed two annotators for 20\% of the data and measured reliability using Cohens’ K 0.71 (95\% CI: [0.55, 0.83]) for Multi-turn and K 0.98 (95\% CI: [0.95, 1.00]) for multilingual PCT dataset. Agreement showed substantial agreement, for more details see Appendix~\ref{sec:evaluation}. 

\subsection{Bias Neuron Identification}
To identify neurons responsible for biased behavior, we attribute the turn-level bias score $S_t$ to internal neuron activations using integrated gradients \cite{yang-etal-2024-mitigating}. For each neuron, we compute an \emph{importance score} that measures its contribution to biased outputs by aggregating attribution strength across tokens and instances. This score is decomposed into two components: $m_{\text{local}}$, capturing the neuron’s relevance in the current turn, and $m_{\text{carry}}$, capturing persistent influence from earlier dialogue turns. This decomposition distinguishes transient, context-specific bias from memory-driven bias propagation, enabling targeted and context-aware mitigation.

\subsection{Aggregation of Bias Scores}
\label{sec:aggregation}
For an instruction $\iota$, an input instance $x_j$ with tokens $\{t_k\}_{k=1}^{K}$, layer $h$, and neuron $i$, we compute per-token attributions $B^{(\iota,x_j,t_k)}_{i}(h)$ (e.g., Grad$\times$Activation). The local and carry-over importance scores ($m_{\text{local}}, m_{\text{carry}}$) are incorporated at this stage by weighting each attribution according to its dialogue-conditioned relevance. We then aggregate over tokens, instances, and instructions to obtain a robust, history-aware score for biased neurons.  
\noindent \textbf{Token aggregation:} Because activations are token-local, we pool over tokens to obtain a single score for $(\iota, x_j)$:

\begin{equation}
B^{(\iota,x_j)}_{i}(h) \;=\; \max_{k \in \{1,\dots,K\}} B^{(\iota,x_j,t_k)}_{i}(h),
\label{eq:token-agg}
\end{equation}

where max-pooling captures the strongest bias-triggering evidence in the input.  

\noindent \textbf{Instance aggregation:} Tasks contain multiple instances $D = \{(x_j,y_j)\}_{j=1}^{N}$. 
We weight instances by a confusion score $\alpha(\iota, x_j)$, scaled by the dialogue-conditioned importance from $m_{\text{local}}$ and $m_{\text{carry}}$, such that persistent carry-over activations are not diluted.  
The aggregated score with normalized weights are:

\begin{equation}
B^{(\iota,D)}_{i}(h) \;=\; \sum_{j=1}^{N} \tilde{\alpha}(\iota,x_j)\, B^{(\iota,x_j)}_{i}(h),
\label{eq:instance-agg}
\end{equation}

\begin{equation}
\tilde{\alpha}(\iota,x_j) = \frac{\alpha(\iota,x_j)}{\sum_{j'=1}^{N} \alpha(\iota,x_{j'})}
\label{eq:confusion}
\end{equation}

where $\alpha(\iota,x_j) = \Pr(\hat{y}_j \mid \iota, x_j)$.

\noindent \textbf{Instruction aggregation:} To be robust across paraphrases, we average over a set of semantically equivalent instructions $I$ (with $|I|=M$):

{\small
\begin{equation}
B^{(I,D)}_{i}(h) \;=\; \frac{1}{M} \sum_{\iota \in I} B^{(\iota,D)}_{i}(h).
\label{eq:instruction-agg}
\end{equation}}

This aggregation provides final rankings $B^{(I,D)}_{i}(h)$ for identifying \emph{bias neurons}, which are then passed to concept tests and adaptive Masking.

\subsection{Concept-Based Testing}
\label{probetest}
To distinguish sudden spikes from persistent bias memory, instead of suppressing all biased neurons, we test which neurons are intact and which have forgotten the skillful knowledge \cite{zhao2024what}. We introduce a \emph{Memory Consistency Probe}
inspired by conditional probing \cite{hewitt2021conditional, belinkov_probing_2022}.  
Contrastive affirmative/negative queries estimate a memory score $C_t$ that quantifies how strongly each neuron preserves biased concepts across the conversation.

\begin{figure*}[ht]
    \centering
    \includegraphics[width=0.9\textwidth]{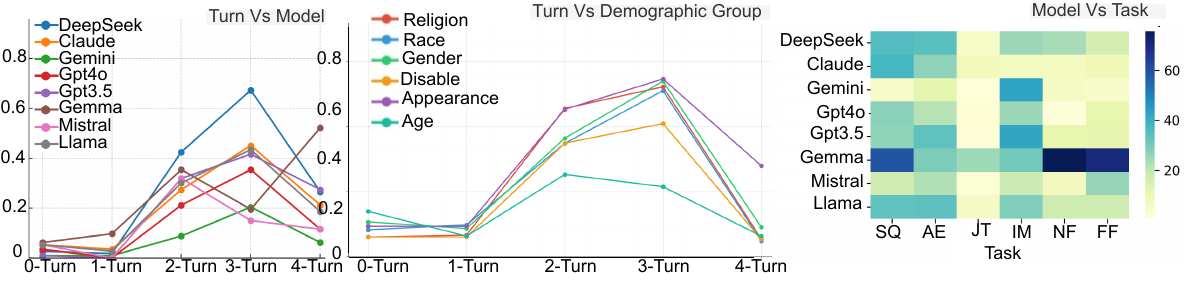}
    \caption{Evaluation of Bias progression across dialogue turns and tasks. (Left) Turn-wise comparison of bias across different models shows an increase in later turns. (Middle) Bias across demographic groups highlights stronger signals for religion, race, and gender in multi-turn contexts.(Right) model task interactions indicate task-specific vulnerability, with Scattered Questions (SQ) and Inference Misinformation (IM) producing higher bias scores across several models.}
    \label{fig:multiTurnEval}
\end{figure*}

\subsection{Dynamic Neuron Masking}
\label{gatingstrength}
Instead of permanently removing neurons, this step adaptively adjusts their influence during generation because bias in conversations changes with context and memory so Dynamic neuron masking is needed to preserve the model's fairness and consistency. Unlike static pruning or fixed neuron masking, which permanently remove neurons irrespective of conversational context. It performs \textit{reversible and graded modulation} of neuron. We apply \emph{controlled masking} that adaptively gates neurons during decoding. For each layer $l$, the gating coefficient is $g_l^{(t)} = \sigma(\alpha S_t + \beta C_t)$, where $\sigma$ is the sigmoid function and $\alpha,\beta$ are tunable weights activations, more details in Appendix~\ref{sec: AdaptiveMASKING}. This mechanism dynamically suppresses high-bias neurons while retaining information essential for coherence and fluency. The proposed Dynamic Neuron Masking framework requires no model retraining and adds only a small per-token cost for scoring and gating, comparable to other real-time steering methods \cite{gupta_advsumm_2025}, algorithm~\ref{alg:dynamic_gating} corresponds to the four-stage pipeline illustrated in Figure~\ref{fig:methodology}. 

\begin{algorithm}[h]
\small
\caption{ Inference Level Mitigation in LLMs}
\label{alg:dynamic_gating}
\begin{algorithmic}[1]

\REQUIRE LLM $f$, dialogue context $\mathbf{x}_{1:t}$, transformer layers $\ell = 1,\dots,L$, instruction set $\mathcal{I}$, evaluation set $\mathcal{D}$, top-$k$ neurons $k$, gating weights $\alpha,\beta$
\ENSURE Bias-mitigated response $\hat{y}_t$

\STATE \textbf{Stage 1: Behavioural Bias Detection}
\STATE $S_t \leftarrow \textsc{BiasScore}(\mathbf{x}_{1:t})$

\STATE \textbf{Stage 2: Bias Neuron Identification}
\FOR{$\ell = 1$ to $L$}
    \STATE Compute attribution scores $A_\ell^{\text{local}}$ from current-turn activations using Gradients
    \STATE Compute attribution scores $A_\ell^{\text{carry}}$ from dialogue-history activations using Gradients
    \STATE $\mathcal{M}_\ell^{\text{local}} \leftarrow \textsc{TopK}(A_\ell^{\text{local}}, k)$ ,  $\mathcal{M}_\ell^{\text{carry}} \leftarrow \textsc{TopK}(A_\ell^{\text{carry}}, k)$
    \STATE $\mathcal{N}_\ell \leftarrow \mathcal{M}_\ell^{\text{local}} \cup \mathcal{M}_\ell^{\text{carry}}$
\ENDFOR

\STATE \textbf{Stage 3: Concept Based Testing}
\STATE $C_t \leftarrow \textsc{MemConsistencyProbe}(\mathbf{x}_{1:t})$

\STATE \textbf{Stage 4: Dynamic Neuron Suppression}, $\mathbf{x}_{1:t}$
\FOR{each decoding step $s$}
    \FOR{$\ell = 1$ to $L$}
        \STATE $g_\ell^{(t)} \leftarrow \sigma(\alpha S_t + \beta C_t)$
        \STATE $\mathbf{h}_\ell[\mathcal{N}_\ell] \leftarrow g_\ell^{(t)} \odot \mathbf{h}_\ell[\mathcal{N}_\ell]$
    \ENDFOR
    \STATE Generate next token using the modified hidden states
\ENDFOR

\STATE $\hat{y}_t \leftarrow f(\mathbf{x}_{1:t}; \{\mathcal{N}_\ell, g_\ell^{(t)}\}_{\ell=1}^{L})$
\STATE \textbf{return} $\hat{y}_t$

\end{algorithmic}
\end{algorithm}

\section{Experiment Settings}
\paragraph{Baseline Comparison:}
Comparisons include comprehensive inference-time baselines based on Prompt Engineering \cite{chisca2024prompting}, logit filtering \cite{fei2023mitigating}, vector steering \cite{siddique_shifting_2025}, and neuron pruning \cite{yang_mitigating_2024}, results in Table~\ref{multiturn-results}. For evaluation we used \href{https://docs.mistral.ai/api/}{Mistral-7B-Instruct-v0.2}-7B and \href{https://lambda.ai}{DeepSeek-Chat}-7B, see more details in Appendix~\ref{appendix:ModelDetails}.
\paragraph{Datasets:}
We evaluate our framework on two complementary benchmarks to capture both single-turn and multi-turn conversational bias.
\textbf{Static and Single Turn:} The multilingual PCT dataset \cite{nadeem2025framing}and benchmark BBQ evaluates social bias in single-turn multiple-choice \cite{parrish_bbq_2022}.

\noindent \textbf{Dynamic and Multi Turn:}
FairMT-Bench \cite{fan_fairmt-bench_nodate} extends bias evaluation to \emph{multi-turn} dialogue and dataset $F^2$Bench is a benchmark that evaluates bias and fairness in LLMs using both single-turn and multi-turn conversational tasks \cite{lan-etal-2025-f2bench}. Bias accumulation and memory carry-over are explicitly annotated, making it ideal for testing context-aware mitigation. Automatic evaluation uses an ensemble: GPT-3.5-Turbo as the primary judge, Llama Guard as the classifier \cite{fan_fairmt-bench_nodate}, and Claude as an auxiliary validator, see details in Appendix~\ref{dataset}.

\paragraph{Hyperparameters} 
We used consistent hyperparameters across all experiments for fairness and reproducibility. Each model was evaluated with a \textit{temperature} = 0.5, and \textit{top-p} = 0.9. The \textit{Dynamic Neuron Masking} framework applied gating coefficients $g_l^{(t)} = \sigma(\alpha S_t + \beta C_t)$, with $\alpha = 1.0$ and $\beta = 0.8$. Bias and memory scores were computed using integrated gradients across all layers, with masking thresholds dynamically adjusted per turn based on dialogue context.

\paragraph{Evaluation Metrics:}
To evaluate the performance the metrics for \textit{Bias Score} is defined as $S_t = \dfrac{n_b}{n_t}$, where $n_b$ is the number of biased texts and $n_t$ is the total number of texts.  
The \textit{Toxicity} metric is $T = \dfrac{n_{tox}}{n_t}$, where $n_{tox}$ denotes the number of toxic texts.  
\textit{Knowledge Retention} is given by $K_t = \dfrac{n_{kr}}{n_t}$, where $n_{kr}$ is the number of texts without knowledge attrition.  
\textit{Faithfulness} is measured as $F = \dfrac{n_{truth}}{n_{claims}}$, with $n_{truth}$ the number of truthful claims and $n_{claims}$ the total number of claims.  
Finally, \textit{Answer Relevancy} is $R = \dfrac{n_{rel}}{n_{stmts}}$, where $n_{rel}$ is the number of relevant statements and $n_{stmts}$ the total number of statements. Together, these metrics provide a balanced protocol: $S_t$ and $T$ quantify fairness and safety, $C_t$ measures factual memory retention, and $F$,$K$  and $R$ ensure responses remain contextually grounded and aligned with user intent to ensure coherence and relevance \cite{deepeval_turnrelevancy}.

\paragraph{Bias Exists:}
We evaluate bias mitigation across political-leaning (PCT) and multi-turn demographic fairness (FairMT) settings. Figure~\ref{fig:multiTurnEval} and Table~\ref{multiturn-results} reports bias scores over both benchmarks and languages. We observe stable gains across turns and topics. These results provide converging evidence that biased neurons exists and mitigating those units reduces biased outputs.

\begin{table}[ht]
\centering
\resizebox{\linewidth}{!}{%
\begin{tabular}{lrrrrr}
\toprule
\textbf{Model} & \textbf{Urdu} & \textbf{Punjabi} & \textbf{Pashto} & \textbf{Sindhi} & \textbf{Balochi} \\
\midrule
GPT-3.5-turbo         & (0.5, -0.1)   & (1.38, 1.95)  & (-0.13, 2.1)   & (1.0, 1.49)   & (1.38, 1.03) \\
GPT-4o                & (-1.75, -1.03) & (-1.5, -2.26) & (-1.13, -0.97) & (0.13, -1.03) & (2.38, 1.08) \\
Claude & (0.25, -1.79)  & (1.13, 0.15)  & (-2.63, -0.26) & (0.0, 0.72)   & (-1.0, 1.59) \\
Gemini           & (-0.75, -2.1)  & (-1.0, 0.31)   & (-0.13, -1.03) & (-0.25, -1.33) & (1.75, 0.77) \\
Mistral & (2.5, 1.23)    & (-1.0, 0.31)   & (0.0, -0.41)   & (-0.75, -2.26) & (1.5, 1.23) \\
DeepSeek           & (-1.0, -1.23)  & (-0.25, -0.05) & (-1.0, 0.87)   & (0.38, -1.28)  & (-2.13, 1.64) \\
Llama                    & (1.88, -0.21)  & (1.63, 0.31)   & (1.38, -0.41)  & (1.63, -0.21)  & (1.75, 0.97) \\
\bottomrule
\end{tabular}%
}
\caption{Evaluation of political inclination on PCT across Languages and Models, value represent stance score on social and economic axis. }
\label{tab:political-compass}
\end{table}

\begin{table*}[ht]
\centering
\small
\resizebox{\textwidth}{!}{
\begin{tabular}{clcccccc}
\hline
\textbf{Model} & 
\textbf{Method evaluated for PCT-SingleTurn} & 
\textbf{English} & 
\textbf{Urdu} & 
\textbf{Punjabi} & 
\textbf{Sindhi} & 
\textbf{Balochi} & 
\textbf{Pashto} \\
\hline
\multirow{6}{*}{\textbf{Mistral}} 
& Baseline & \textcolor{black}{46.7 (±2.1)} & \textcolor{black}{49.4 (±2.4)} & \textcolor{black}{57.7 (±2.7)} & \textcolor{black}{65.8 (±3.0)} & \textcolor{black}{54.5 (±2.5)} & \textcolor{black}{75.8 (±3.2)} \\
& Prompt Engineering & 30.6 (±1.8) & 28.1 (±1.5) & 46.7 (±2.2) & 64.7 (±2.7) & 53.1 (±2.1) & 66.4 (±2.5) \\
& Output Filtering & 28.5 (±1.6) & 26.3 (±1.4) & 34.9 (±1.9) & 58.2 (±2.6) & 51.7 (±2.0) & 67.1 (±2.4) \\
& Steering Vectors Ensembles & 24.8 (±1.5) & 31.6 (±1.8) & 51.2 (±2.3) & 43.4 (±2.0) & 48.6 (±2.1) & 66.8 (±2.5) \\
& Neuron Pruning & 21.7 (±1.2) & 26.8 (±1.6) & 40.5 (±2.0) & 41.1 (±1.9) & 46.9 (±2.1) & 54.3 (±2.2) \\
& \textbf{Dynamic Neurons Masking (Proposed)} & \textbf{17.3 (±1.1)} & \textbf{21.2 (±1.3)} & \textbf{29.1 (±1.6)} & \textbf{24.1 (±1.4)} & \textbf{22.5 (±1.3)} & \textbf{31.0 (±1.5)} \\
\hline
\multirow{6}{*}{\textbf{DeepSeek}} 
& Baseline & \textcolor{black}{33.2 (±2.0)} & \textcolor{black}{37.0 (±2.1)} & \textcolor{black}{47.3 (±2.4)} & \textcolor{black}{45.2 (±2.3)} & \textcolor{black}{64.1 (±2.7)} & \textcolor{black}{65.4 (±2.8)} \\
& Prompt Engineering & 30.3 (±1.7) & 28.0 (±1.5) & 46.5 (±2.1) & 51.2 (±2.2) & 56.9 (±2.3) & 57.3 (±2.4) \\
& Output Filtering & 29.1 (±1.6) & 29.5 (±1.7) & 45.3 (±2.2) & 54.4 (±2.3) & 44.5 (±2.0) & 44.3 (±2.1) \\
& Steering Vectors Ensembles & 23.0 (±1.5) & 20.7 (±1.3) & 33.9 (±1.8) & 52.7 (±2.2) & 42.3 (±2.0) & 59.5 (±2.4) \\
& Neuron Pruning & 12.5 (±1.0) & 24.1 (±1.5) & 32.1 (±1.8) & 41.3 (±2.1) & 41.2 (±2.0) & 47.9 (±2.2) \\
& \textbf{Dynamic Neurons Masking (Proposed)} & \textbf{11.2 (±0.9)} & \textbf{14.6 (±1.1)} & \textbf{19.0 (±1.3)} & \textbf{22.9 (±1.5)} & \textbf{20.2 (±1.2)} & \textbf{21.0 (±1.4)} \\
\midrule
\textbf{Model} & 
\textbf{Method evaluated for BBQ-SingleTurn} & 
\textbf{Age} & 
\textbf{Disability} & 
\textbf{Race} & 
\textbf{Appearance} & 
\textbf{Gender} & 
\textbf{Religion} \\
\hline
\multirow{6}{*}{\textbf{Mistral}} 
& Baseline 
& \textcolor{black}{34.8 (±1.0)} & \textcolor{black}{27.6 (±2.3)} & \textcolor{black}{45.9 (±2.6)} & \textcolor{black}{33.2 (±2.9)} & \textcolor{black}{42.8 (±2.4)} & \textcolor{black}{53.1 (±3.1)} \\
& Prompt Engineering & 34.5 (±1.8) & 32.9 (±1.9) & 38.7 (±2.3) & 38.4 (±2.7) & 29.6 (±2.5) & 25.8 (±2.6) \\
& Output Filtering & 31.7 (±1.6) & 23.2 (±1.7) & 32.1 (±2.1) & 24.6 (±1.5) & 36.8 (±2.4) & 41.9 (±2.4) \\
& Steering Vectors Ensembles & 28.9 (±1.5) & 21.0 (±1.6) & 19.8 (±1.0) & 38.2 (±1.2) & 34.1 (±1.0) & 38.6 (±2.3) \\
& Neuron Pruning & 25.4 (±1.3) & 27.9 (±1.5) & 35.6 (±1.9) & 24.3 (±2.1) & 21.5 (±2.9) & 32.7 (±2.2) \\
& \textbf{Dynamic Neurons Masking (Proposed)} & \textbf{19.6 (±1.1)} & \textbf{22.8 (±1.3)} & \textbf{28.7 (±1.6)} & \textbf{20.9 (±1.5)} & \textbf{26.4 (±1.4)} & \textbf{21.2 (±1.6)} \\
\hline
\multirow{6}{*}{\textbf{DeepSeek}} 
& Baseline & \textcolor{black}{26.5 (±1.9)} & \textcolor{black}{33.8 (±2.1)} & \textcolor{black}{39.2 (±2.4)} & \textcolor{black}{36.8 (±2.3)} & \textcolor{black}{40.9 (±2.6)} & \textcolor{black}{33.7 (±2.7)} \\
& Prompt Engineering & 21.2 (±1.7) & 24.5 (±1.8) & 35.0 (±1.1) & 23.6 (±1.0) & 36.2 (±2.3) & 28.4 (±2.4) \\
& Output Filtering & 29.4 (±1.6) & 32.1 (±1.7) & 21.7 (±2.0) & 41.9 (±2.0) & 30.8 (±2.1) & 32.6 (±2.2) \\
& Steering Vectors Ensembles & 25.8 (±1.5) & 27.4 (±1.4) & 25.1 (±1.8) & 39.2 (±1.9) & 45.7 (±2.0) & 48.9 (±2.2) \\
& Neuron Pruning & 22.1 (±1.3) & 24.8 (±1.5) & 31.9 (±1.7) & 36.1 (±1.8) & 42.6 (±1.9) & 44.5 (±2.1) \\
& \textbf{Dynamic Neurons Masking (Proposed)} 
& \textbf{16.3 (±1.0)} & \textbf{18.9 (±1.1)} & \textbf{19.7 (±1.3)} & \textbf{27.6 (±1.4)} & \textbf{24.1 (±1.3)} & \textbf{21.8 (±1.4)} \\
\midrule
\textbf{Model} & \textbf{Method evaluated for$F^2$-Bench} & 
\textbf{Age} & 
\textbf{Nationality} & 
\textbf{Race} & 
\textbf{Appearance} & 
\textbf{Gender} & 
\textbf{Religion} \\
\hline
\multirow{6}{*}{\textbf{Mistral}} 
& Baseline & 43.1 (±1.4) & 59.8 (±1.6) & 48.7 (±1.0) & 57.2 (±1.4) & 54.1 (±1.5) 
& 47.36 (±1.4) \\
& Prompt Engineering & 41.8 (±1.3) & 44.7 (±1.5) & 47.3 (±1.9) & 42.6 (±1.8) & 43.8 (±1.33) & 38.29 (±1.3) \\
& Output Filtering & 40.1 (±1.2) & 41.5 (±2.1) & 38.3 (±1.4) & 41.4 (±2.1) & 39.5 (±1.4) & 34.62 (±1.0) \\
& Steering Vectors Ensembles & 32.4 (±2.3) & 34.6 (±2.0) & 31.5 (±2.0) & 37.7 (±2.0) & 31.9 (±2.2) & 30.8 (±2.0) \\
& Neuron Pruning & 27.5 (±1.8) & 29.0 (±2.0) & 25.1 (±1.9) &29.9 (±2.1) & 29.8 (±2.0) & 25.9 (±2.0) \\
& \textbf{Dynamic Neurons Masking (Proposed)} & \textbf{21.0 (±1.0)} & \textbf{20.8 (±1.6)} & \textbf{22.0 (±1.7)} & \textbf{17.4 (±1.5)} & \textbf{19.9 (±1.3)} & \textbf{18.6 (±1.5)} \\
\hline

\multirow{6}{*}{\textbf{DeepSeek}} 
& Baseline & \textcolor{black}{40.2 (±1.91)} & \textcolor{black}{33.0 (±1.67)} & \textcolor{black}{45.7 (±2.1)} & \textcolor{black}{31.7 (±1.5)} & \textcolor{black}{43.6 (±1.4)} & \textcolor{black}{42.87 (±1.3)} \\
& Prompt Engineering & 38.6 (±2.6) & 30.7 (±1.6) & 41.1 (±1.9) & 30.2 (±2.3) & 39.2 (±2.1) & 37.7 (±1.6) \\
& Output Filtering & 33.2 (±2.0) & 30.2 (±2.0) & 39.4 (±2.5) & 27.5 (±2.0) & 34.3 (±2.0) & 36.1 (±2.4) \\
& Steering Vectors Ensembles & 21.4 (±2.3) & 26.7 (±2.1) & 36.3 (±1.7) & 26.9 (±1.9) & 31.1 (±2.0) & 29.2 (±2.0)  \\
& Neuron Pruning & 19.3 (±2.0) & 15.8 (±1.0) & 18.9 (±2.2) & 21.6 (±1.0) & 23.8 (±2.2) & 25.7 (±1.6) \\
& \textbf{Dynamic Neurons Masking (Proposed)} & \textbf{18.7 (±1.8)} & \textbf{12.9 (±1.2)} & \textbf{19.8 (±1.2)} & \textbf{19.0 (±2.0)} & \textbf{17.9 (±1.6)} & \textbf{15.8 (±1.2)} \\
\midrule
\textbf{Model} & \textbf{Method evaluated for FairMT-Bench} & 
\textbf{Anaphora Elipsis} & 
\textbf{Scattered Questions} & 
\textbf{Jailbreak Tips} & 
\textbf{Inte. Misinfo} & 
\textbf{Negative Feedback} & 
\textbf{Fixed Format} \\
\hline
\multirow{6}{*}{\textbf{Mistral}} 
& Baseline & \textcolor{black}{42.1 (±2.0)} & \textcolor{black}{57.1 (±2.3)} & \textcolor{black}{22.5 (±1.2)} & \textcolor{black}{38.2 (±1.9)} & \textcolor{black}{67.0 (±2.6)} & \textcolor{black}{35.8 (±1.8)} \\
& Prompt Engineering & 30.3 (±1.5) & 43.5 (±1.9) & 35.2 (±1.6) & 20.9 (±1.3) & 50.2 (±2.0) & 33.4 (±1.7) \\
& Output Filtering & 47.4 (±2.1) & 42.0 (±1.8) & 34.1 (±1.5) & 39.6 (±1.9) & 48.7 (±2.1) & 29.9 (±1.5) \\
& Steering Vectors Ensembles & 32.3 (±1.7) & 35.1 (±1.6) & 42.5 (±1.9) & 36.7 (±1.8) & 45.2 (±1.9) & 23.1 (±1.3) \\
& Neuron Pruning & 37.0 (±1.8) & 43.4 (±1.9) & 22.0 (±1.2) & 32.1 (±1.7) & 35.8 (±1.6) & 25.5 (±1.4) \\
& \textbf{Dynamic Neurons Masking (Proposed)} & \textbf{21.1 (±1.3)} & \textbf{32.7 (±1.6)} & \textbf{29.4 (±1.5)} & \textbf{27.3 (±1.4)} & \textbf{19.1 (±1.2)} & \textbf{8.5 (±0.8)} \\
\hline
\multirow{6}{*}{\textbf{DeepSeek}} 
& Baseline & \textcolor{black}{34.2 (±1.9)} & \textcolor{black}{45.3 (±2.0)} & \textcolor{black}{25.0 (±1.3)} & \textcolor{black}{45.1 (±2.1)} & \textcolor{black}{43.3 (±2.0)} & \textcolor{black}{26.1 (±1.5)} \\
& Prompt Engineering & 31.5 (±1.7) & 41.4 (±1.8) & 33.0 (±1.5) & 44.1 (±2.0) & 42.3 (±1.9) & 26.5 (±1.6) \\
& Output Filtering & 33.1 (±1.6) & 38.3 (±1.7) & 25.5 (±1.3) & 39.8 (±1.9) & 43.1 (±2.0) & 31.0 (±1.7) \\
& Steering Vectors Ensembles & 27.1 (±1.5) & 38.8 (±1.8) & 23.3 (±1.2) & 37.8 (±1.8) & 31.1 (±1.5) & 33.3 (±1.6) \\
& Neuron Pruning & 26.3 (±1.5) & 32.6 (±1.6) & 20.6 (±1.1) & 39.7 (±1.9) & 34.3 (±1.7) & 23.5 (±1.4) \\
& \textbf{Dynamic Neurons Masking (Proposed)} & \textbf{10.7 (±0.9)} & \textbf{24.9 (±1.3)} & \textbf{16.0 (±1.0)} & \textbf{19.9 (±1.2)} & \textbf{22.3 (±1.4)} & \textbf{29.1 (±1.3)} \\
\hline
\end{tabular}}
\caption{Comparison of bias mitigation methods (lower is better). Values in parentheses denote 95\% bootstrap CI. The baseline shows higher bias due to the absence of representation-level intervention, while static methods reduce bias but may degrade coherence and utility. In contrast, the proposed method achieves stronger bias reduction through dynamic neuron masking, suppressing bias-related activations while preserving response quality.}
\label{multiturn-results}
\end{table*}

\begin{figure}[htbp]
    \includegraphics[width=0.46\textwidth]{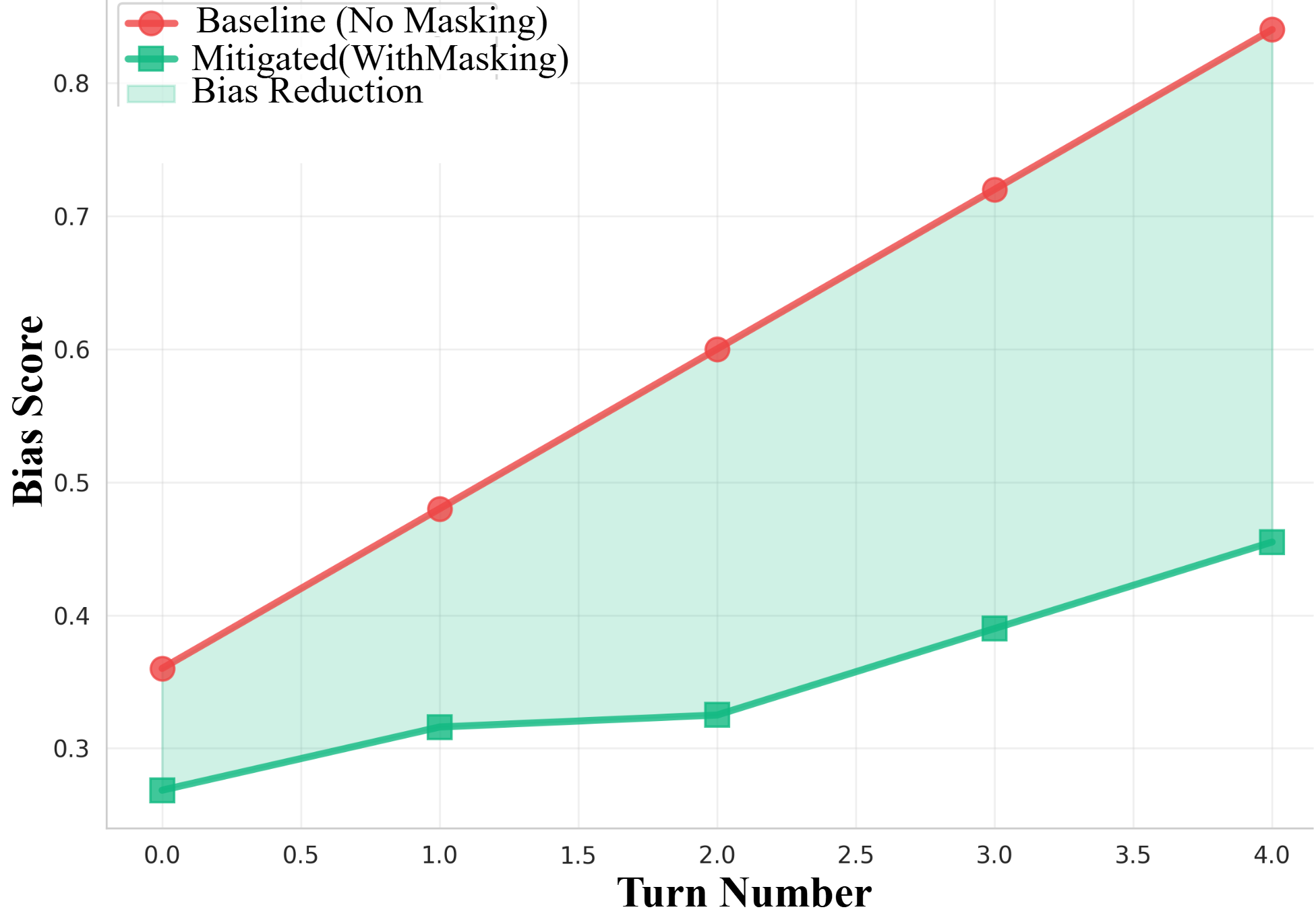} 
    \caption{Mitigation comparison of effective bias scores across dialogue turns, baseline (red, no masking) and mitigation (green, with masking) showing higher bias accumulation in the baseline (no masking) and substantial reduction with masking applied.}
    \label{fig:baseline_vs_mitigated}
\end{figure}

\section{Result and Analysis}

In the Table~\ref{multiturn-results}, the results show without any intervention, both Mistral and DeepSeek carry a high level of bias across languages and dialogue tasks. Simple fixes like prompt engineering or output filtering help a little but don’t fully solve the problem, while methods such as steering vectors ensembles, and pruning improve things but often lost semantics and coherence with dialogue relevance. 

Our proposed Dynamic Neurons Masking stands out by consistently bringing the bias scores down across all languages in PCT and BBQ and across all task types in FairMT, and $F^2$ bench in MultiTurn. Our framework not only reduces bias more effectively but also works reliably in complex, real dialogue situations. In the baseline, the effective bias score rises steadily with each turn, indicating that bias accumulates as the conversation continues. 

\begin{figure}[h]
    \centering
    \includegraphics[width=0.45\textwidth]{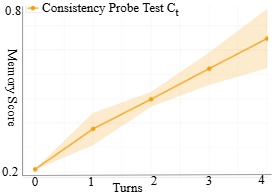} 
    \caption{Probe Testing for model Memory Score increases steadily across turns, indicating stronger memory retention in later dialogue, which enhances consistency but also raises the risk of bias carry-over.}
    \label{fig:memoryscore}
\end{figure}

Whereas, Table~\ref{multiturn-results} demonstrates consistent bias reduction across all evaluated languages, with the largest improvements in challenging cases such as Sindhi and Pashto, resulting in more stable and reliable responses across multilingual settings. In contrast, the mitigated curve grows more slowly, demonstrating that masking effectively dampens bias propagation across turns. The shaded region represents the extent of bias reduction achieved by masking, clearly showing that the method consistently lowers bias throughout the dialogue. This highlights the effectiveness of masking in reducing bias carry-over while maintaining dialogue progression as shown in Figure~\ref{fig:baseline_vs_mitigated}. This illustrates how the model learned and remembered and how the knowledge remained intact; the model remembers knowledge that evolves over successive dialogue turns. Starting from a relatively low value at the first turn, the score rises steadily with each additional turn, showing that the model retains more information from previous context as the conversation progresses. The shaded confidence interval indicates that while there is some variation across conversations, the overall trend is consistent: memory carry-over strengthens over time. In Figure~\ref{fig:memoryscore}, the pattern highlights both the benefit of improved contextual consistency and the potential drawback of increased bias persistence in longer dialogues.

\begin{figure}[ht]
    \includegraphics[width=0.5\textwidth]{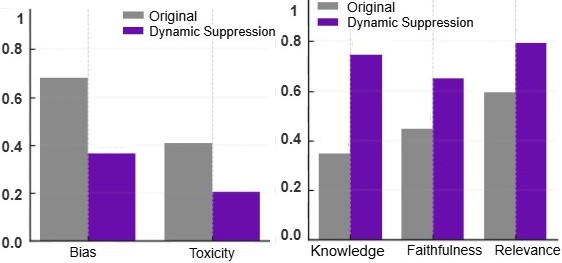} 
    \caption{Dynamic masking reduces stereotype bias and toxicity while improving knowledge accuracy, faithfulness, and relevance.}
    \label{fig:coherence}
\end{figure}

Dynamic Masking substantially reduces stereotype bias and toxicity while simultaneously improving knowledge accuracy, faithfulness, and relevance, demonstrating a favorable bias utility trade-off, as  shown in Figure~\ref{fig:coherence}. In multi-turn settings, the DeepSeek model in its original inference mode exhibited bias carry-over and contextual drift, particularly on anaphora ellipsis tasks, where biased activations propagated across turns and degraded coherence and factual consistency. Applying Dynamic Neuron Suppression filters these biased traces before they accumulate in dialogue memory, enabling the model to maintain coherent context, accurate references, and faithful responses. Table~\ref{MaskedResponseResults} further shows that dynamic neuron gating produces more balanced, neutral, and context-aware responses without sacrificing coherence. For generalization to Larger Instruction-Tuned Model, we further evaluate the proposed pipeline using the Mistral-Small-3.2-24B-Instruct-2506 model\footnote{\url{https://huggingface.co/mistralai/Mistral-Small-3.2-24B-Instruct-2506}}, 
an instruction-tuned language model on multi-turn dataset, the baseline model shows bias as, age:38.29, gender:33.04, and appearance: 31.72. After applying Dynamic Neuron Masking, performance improves to age:23.38, gender:26.42, and appearance:21.95, outperforming prompt engineering, output filtering, and steering-based baselines.

\begin{figure}[ht]
    \includegraphics[width=0.45\textwidth]{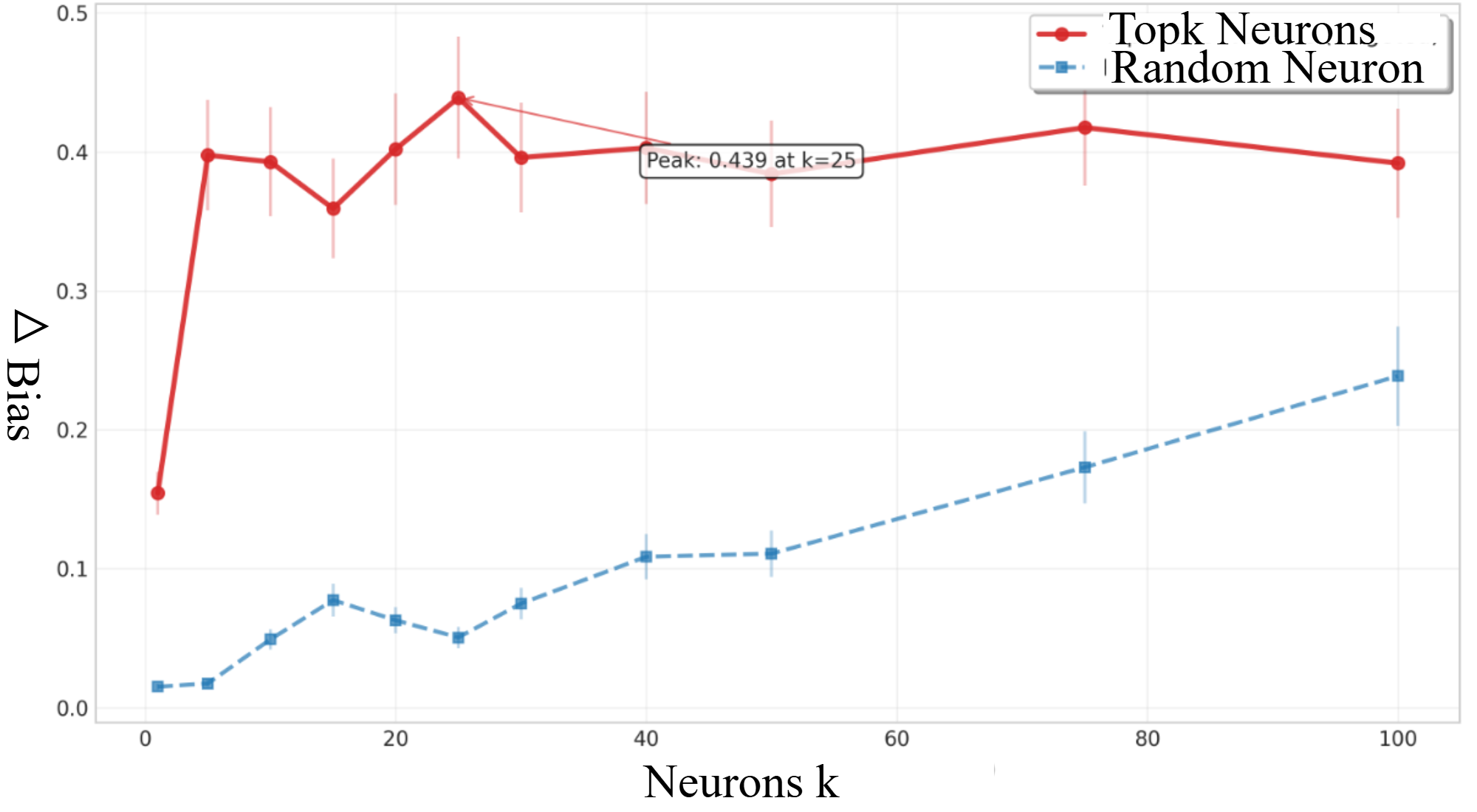} 
    \caption{Bias attribution strength across neurons. Top-k selected neurons (red) consistently capture higher bias signals compared to randomly chosen neurons (blue), demonstrating that targeted attribution isolates bias-relevant activations more effectively. Error bars indicate standard deviation across runs.}
    \label{fig:neuron-ablation}
\end{figure}

\subsection{Ablation Studies}
We perform ablation experiments to isolate the contribution of each component in the proposed Dynamic Neuron Suppression framework and to explain why static or partial interventions fail in multi-turn dialogue. \textit{Static masking without memory probes,} removing the memory consistency probe (Section~\ref{probetest}) and applying a fixed neuron mask across turns reduces bias in early turns but degrades performance as dialogue progresses, as shown in Table~\ref{tab:multi_turn_utility}. We observe higher refusal rates, reduced relevance, and topic drift, indicating that bias re-emerges through memory carry-over when masking is not refreshed in a context-aware manner. \textit{Unconditional neuron removal (static pruning),}  
ablating the gating mechanism (Section~\ref{gatingstrength}) and permanently zeroing out biased neurons yields the largest immediate bias reduction, but severely harms utility: responses become shorter, less specific, and off-topic, even for neutral inputs. This confirms that bias-related neurons also support general semantic and factual reasoning. Figure~\ref{fig:neuron-ablation} shows that bias is localized in a small subset of neurons: targeted ablation achieves a peak $\Delta$Bias of 0.439 at $k=25$, while random ablation yields only marginal gains. 
\begin{table}[htbp] 
\centering 
\small 
\setlength{\tabcolsep}{1pt} 
\begin{tabular}{lccccc} 
\hline 
\textbf{Stage} & \textbf{Knowledge } & \textbf{Fluency} & \textbf{Coherence} & \textbf{Faithfulness} \\ 
\hline 
Skip Attr. & 0.75 & 0.82 & 0.74 & 0.72 \\ 
Skip Probe & 0.77 & 0.80 & 0.73 & 0.71 \\ 
Static Prune & 0.61 & 0.71 & 0.65 & 0.60 \\ 
Proposed & \textbf{0.93} & \textbf{0.85} & \textbf{0.90} & \textbf{0.88} \\ 
\hline 
\end{tabular}
\caption{Ablation studies: Skipping each stage showing utility degradation, where Static pruning provides largest drop in knowledge and coherence.} 
\label{tab:multi_turn_utility} 
\end{table}
Table~\ref{tab:multi_turn_utility} shows that skipping any stage degrades either fairness or core language abilities, while the full pipeline achieves the best bias utility trade-off, underscoring the necessity of dynamic, memory-aware mitigation.
\section{Conclusion}
Bias in LLMs is not a fixed defect; it accumulates across turns. We present dynamic neuron masking, an inference-time approach that identifies, attributes, and selectively gates bias-carrying activations as dialogue unfolds. Evaluated on multilingual PCT and FairMT-Bench, it consistently reduces bias while maintaining fluency, faithfulness, and relevance, outperforming prompt edits, logit filters, steering, and static pruning. Ablations show memory probes prevent rebound, and targeted gating avoids over-suppression, enabling reversible fairness.
\section*{Limitations}
While our framework demonstrates strong improvements across multilingual and multi-turn benchmarks, several limitations remain. Our evaluation primarily focuses on fairness, toxicity and demographic stereotypes; extending the analysis to broader domains such as healthcare, education, legal decision-making, and other multilingualism or culturally sensitive applications remains an important direction for future work. Although the proposed neuron-level masking mechanism is computationally streamlined, it introduces additional overhead during decoding, which may pose scalability challenges for very large models or high-throughput deployments. Furthermore, human evaluation was conducted on selected subsets of the data, leaving open questions about the generalization of the findings across unseen contexts, domains, and real-world applications.
\section*{Ethical Concerns}
This work directly engages with fairness and bias mitigation in LLMs, which carries both opportunities and risks. Our approach reduces harmful stereotypes and toxic carry-over, but there is no guarantee of complete neutrality: Masking could inadvertently mask legitimate perspectives or reduce transparency about model behavior. Misuse is possible if such techniques are applied to obscure rather than reveal bias. Moreover, our benchmarks focus on specific languages and demographics; excluding others may inadvertently reinforce blind spots. Ethical deployment requires complementing our method with human oversight, continuous auditing, and culturally grounded evaluations to ensure inclusivity and accountability.

\bibliography{custom}

\begin{thebibliography}{55}
\providecommand{\natexlab}[1]{#1}

\bibitem[{Abbas and Bidin(2022)}]{abbas2022languageidentity}
Fozia Abbas and Saadiyah~Javed Bidin. 2022.
\newblock Language policy and planning in pakistan: A critical analysis of the language planning and policy (lpp) in pakistan and its impact on indigenous languages of pakistan.
\newblock \emph{Eurasian Journal of Applied Linguistics}, pages 85--96.

\bibitem[{Adewumi et~al.(2024)Adewumi, Alkhaled, Gurung, van Boven, and Pagliai}]{adewumi2024fairness}
Tosin Adewumi, Lama Alkhaled, Namrata Gurung, Goya van Boven, and Irene Pagliai. 2024.
\newblock \href {https://doi.org/10.48550/arXiv.2406.19097} {Fairness and bias in multimodal {AI}: A survey}.
\newblock arXiv preprint arXiv:2406.19097.
\newblock Last revised 7 September 2024.

\bibitem[{An et~al.(2024)An, Acquaye, Wang, Li, and Rudinger}]{an_large_2024}
Haozhe An, Christabel Acquaye, Colin Wang, Zongxia Li, and Rachel Rudinger. 2024.
\newblock \href {https://doi.org/10.18653/v1/2024.acl-short.37} {Do {Large} {Language} {Models} {Discriminate} in {Hiring} {Decisions} on the {Basis} of {Race}, {Ethnicity}, and {Gender}?}
\newblock In \emph{Proceedings of the 62nd {Annual} {Meeting} of the {Association} for {Computational} {Linguistics} ({Volume} 2: {Short} {Papers})}, pages 386--397, Bangkok, Thailand. Association for Computational Linguistics.

\bibitem[{Bayasi et~al.(2024)Bayasi, Fayyad, Hamarneh, Garbi, and Najjaran}]{bayasi_debiasify_nodate}
Nourhan Bayasi, Jamil Fayyad, Ghassan Hamarneh, Rafeef Garbi, and Homayoun Najjaran. 2024.
\newblock Debiasify: Self-distillation for unsupervised bias mitigation.

\bibitem[{Behnke and Heafield(2020)}]{behnke-heafield-2020-losing}
Maximiliana Behnke and Kenneth Heafield. 2020.
\newblock \href {https://doi.org/10.18653/v1/2020.emnlp-main.211} {Losing heads in the lottery: Pruning transformer attention in neural machine translation}.
\newblock In \emph{Proceedings of the 2020 Conference on Empirical Methods in Natural Language Processing (EMNLP)}, pages 2664--2674, Online. Association for Computational Linguistics.

\bibitem[{Belinkov(2022)}]{belinkov_probing_2022}
Yonatan Belinkov. 2022.
\newblock \href {https://doi.org/10.1162/coli_a_00422} {Probing classifiers: Promises, shortcomings, and advances}.
\newblock 48(1):207--219.

\bibitem[{Bender and Friedman(2018)}]{bender_data_2018}
Emily~M. Bender and Batya Friedman. 2018.
\newblock \href {https://doi.org/10.1162/tacl_a_00041} {Data statements for natural language processing: Toward mitigating system bias and enabling better science}.
\newblock 6:587--604.

\bibitem[{Bouchard et~al.(2025)Bouchard, Chauhan, Skarbrevik, Bajaj, and Ahmad}]{bouchard_langfair_2025}
Dylan Bouchard, Mohit~Singh Chauhan, David Skarbrevik, Viren Bajaj, and Zeya Ahmad. 2025.
\newblock \href {https://doi.org/10.21105/joss.07570} {{LangFair}: A python package for assessing bias and fairness in large language model use cases}.
\newblock 10(105):7570.

\bibitem[{Chen et~al.(2025)Chen, Wang, Xue, Gao, and Zhou}]{chen2025steering}
Jing Chen, Liyuan Wang, Xiang Xue, Shuo Gao, and Xiaojun Zhou. 2025.
\newblock Steering large language models for safety and robustness.
\newblock \emph{arXiv preprint arXiv:2508.08846}.

\bibitem[{Chisca et~al.(2024)Chisca, Rad, and Lemnaru}]{chisca2024prompting}
Andrei-Victor Chisca, Andrei-Cristian Rad, and Camelia Lemnaru. 2024.
\newblock Prompting fairness: Learning prompts for debiasing large language models.
\newblock In \emph{Proceedings of the Fourth Workshop on Language Technology for Equality, Diversity, Inclusion (LTEDI 2024)}, pages 52--62.

\bibitem[{Dai et~al.(2025)Dai, Xu, Xu, Pang, Dong, and Xu}]{dai_unifying_2025}
Sunhao Dai, Chen Xu, Shicheng Xu, Liang Pang, Zhenhua Dong, and Jun Xu. 2025.
\newblock \href {https://doi.org/10.1145/3701551.3703478} {Unifying bias and unfairness in information retrieval: New challenges in the {LLM} era}.
\newblock In \emph{Proceedings of the Eighteenth {ACM} International Conference on Web Search and Data Mining}, pages 998--1001. {ACM}.

\bibitem[{{DeepEval}(2025)}]{deepeval_turnrelevancy}
{DeepEval}. 2025.
\newblock Turn relevancy | deepeval: The open-source llm evaluation framework.
\newblock \url{https://deepeval.com/docs/metrics-turn-relevancy}.
\newblock Accessed: 2025-09-20.

\bibitem[{Defrance et~al.(2025)Defrance, Bied, Buyl, Lijffijt, and Bie}]{defrance_bimi_2025}
{MaryBeth} Defrance, Guillaume Bied, Maarten Buyl, Jefrey Lijffijt, and Tijl~De Bie. 2025.
\newblock \href {https://doi.org/10.48550/arXiv.2505.22114} {{BiMi} sheets: Infosheets for bias mitigation methods}.
\newblock \emph{Preprint}, arxiv:2505.22114 [cs].

\bibitem[{Demszky et~al.(2019)Demszky, Garg, Voigt, Zou, Shapiro, Gentzkow, and Jurafsky}]{demszky_analyzing_2019}
Dorottya Demszky, Nikhil Garg, Rob Voigt, James Zou, Jesse Shapiro, Matthew Gentzkow, and Dan Jurafsky. 2019.
\newblock \href {https://doi.org/10.18653/v1/N19-1304} {Analyzing {Polarization} in {Social} {Media}: {Method} and {Application} to {Tweets} on 21 {Mass} {Shootings}}.
\newblock In \emph{Proceedings of the 2019 {Conference} of the {North}}, pages 2970--3005, Minneapolis, Minnesota. Association for Computational Linguistics.

\bibitem[{Dufort-Labbé et~al.(2024)Dufort-Labbé, D’Oro, Nikishin, Pascanu, Bacon, and Baratin}]{Maxwelldemon}
Simon Dufort-Labbé, Pierluca D’Oro, Evgenii Nikishin, Razvan Pascanu, Pierre-Luc Bacon, and Aristide Baratin. 2024.
\newblock \href {https://arxiv.org/abs/2403.07688} {Maxwell’s demon at work: Efficient pruning by leveraging saturation of neurons}.
\newblock \emph{Preprint}, arXiv:2403.07688.

\bibitem[{Fan et~al.(2024)Fan, Chen, Hu, and Liu}]{fan_fairmt-bench_nodate}
Zhiting Fan, Ruizhe Chen, Tianxiang Hu, and Zuozhu Liu. 2024.
\newblock {FAIRMT}-{BENCH}: {BENCHMARKING} {FAIRNESS} {FOR} {MULTI}-{TURN} {DIALOGUE} {IN} {CONVERSATIONAL} {LLMS}.

\bibitem[{Fei et~al.(2023)Fei, Hou, Chen, and Bosselut}]{fei2023mitigating}
Yu~Fei, Yifan Hou, Zeming Chen, and Antoine Bosselut. 2023.
\newblock \href {https://doi.org/10.48550/arXiv.2305.19148} {Mitigating label biases for in-context learning}.
\newblock \emph{arXiv preprint arXiv:2305.19148}.

\bibitem[{Geva et~al.(2021)Geva, Schuster, Berant, and Levy}]{geva_transformer_2021}
Mor Geva, Roei Schuster, Jonathan Berant, and Omer Levy. 2021.
\newblock \href {https://doi.org/10.18653/v1/2021.emnlp-main.446} {Transformer feed-forward layers are key-value memories}.
\newblock In \emph{Proceedings of the 2021 Conference on Empirical Methods in Natural Language Processing}, pages 5484--5495. Association for Computational Linguistics.

\bibitem[{Gupta et~al.(2024)Gupta, Sethi, and Sethi}]{gupta2024notraining}
Aviral Gupta, Armaan Sethi, and Ameesh Sethi. 2024.
\newblock No training wheels: Steering vectors for bias correction at inference time.
\newblock In \emph{Proceedings of the 41st International Conference on Machine Learning (ICML)}.
\newblock *Equal contribution.

\bibitem[{Gupta et~al.(2025)Gupta, Varimalla, Deas, Subbiah, and {McKeown}}]{gupta_advsumm_2025}
Mukur Gupta, Nikhil~Reddy Varimalla, Nicholas Deas, Melanie Subbiah, and Kathleen {McKeown}. 2025.
\newblock \href {https://doi.org/10.48550/arXiv.2506.06273} {{AdvSumm}: Adversarial training for bias mitigation in text summarization}.
\newblock \emph{Preprint}, arxiv:2506.06273 [cs].

\bibitem[{Helwe et~al.(2025)Helwe, Balalau, and Ceolin}]{helwe2025multilingualpct}
Chadi Helwe, Oana Balalau, and Davide Ceolin. 2025.
\newblock Navigating the political compass: Evaluating multilingual llms across languages and nationalities.
\newblock In \emph{Findings of the Association for Computational Linguistics: ACL 2025}, pages 17179--17204, Bangkok, Thailand. Association for Computational Linguistics.

\bibitem[{Hewitt et~al.(2021)Hewitt, Ethayarajh, Liang, and Manning}]{hewitt2021conditional}
John Hewitt, Kawin Ethayarajh, Percy Liang, and Christopher~D. Manning. 2021.
\newblock \href {https://arxiv.org/pdf/2109.09234} {Conditional probing: Measuring usable information beyond a baseline}.
\newblock \emph{arXiv preprint arXiv:2109.09234}.

\bibitem[{Hua et~al.(2023)Hua, Ge, Xu, Ji, and Zhang}]{hua_up5_nodate}
Wenyue Hua, Yingqiang Ge, Shuyuan Xu, Jianchao Ji, and Yongfeng Zhang. 2023.
\newblock {UP}5: Unbiased foundation model for fairness-aware recommendation.

\bibitem[{Ibrahim et~al.(2025)Ibrahim, Akbulut, Elasmar, Rastogi, Kahng, Morris, McKee, Rieser, Shanahan, and Weidinger}]{ibrahim_multi-turn_2025}
Lujain Ibrahim, Canfer Akbulut, Rasmi Elasmar, Charvi Rastogi, Minsuk Kahng, Meredith~Ringel Morris, Kevin~R. McKee, Verena Rieser, Murray Shanahan, and Laura Weidinger. 2025.
\newblock \href {https://doi.org/10.48550/arXiv.2502.07077} {Multi-turn evaluation of anthropomorphic behaviours in large language models}.
\newblock arXiv preprint arXiv:2502.07077.
\newblock Submitted on 10 February 2025.

\bibitem[{Li et~al.(2025)Li, Shen, Yao, Ding, Miao, Krishnan, and Padman}]{li2025beyond}
Yubo Li, Xiaobin Shen, Xinyu Yao, Xueying Ding, Yidi Miao, Ramayya Krishnan, and Rema Padman. 2025.
\newblock Beyond single-turn: A survey on multi-turn interactions with large language models.
\newblock \emph{arXiv preprint arXiv:2504.04717}.

\bibitem[{Lin et~al.(2020)Lin, Liu, Yang, Hua, and Roth}]{lin2020pruning}
Zi~Lin, Jeremiah~Zhe Liu, Zi~Yang, Nan Hua, and Dan Roth. 2020.
\newblock \href {https://arxiv.org/abs/2010.01791} {Pruning redundant mappings in transformer models via spectral-normalized identity prior}.
\newblock \emph{arXiv preprint arXiv:2010.01791}.

\bibitem[{Liu et~al.(2023)Liu, Li, Jiang, Ni, Hu, and Huang}]{liu2023towards}
Xinyang Liu, Peilin Li, Hui Jiang, Jun Ni, Xiaodong Hu, and Xiaolei Huang. 2023.
\newblock Towards understanding and mitigating bias in large language models.
\newblock \emph{arXiv preprint arXiv:2305.19148}.

\bibitem[{Ma et~al.(2023)Ma, Fang, and Wang}]{ma2023llmpruner}
Xinyin Ma, Gongfan Fang, and Xinchao Wang. 2023.
\newblock \href {https://arxiv.org/abs/2305.11627} {Llm-pruner: On the structural pruning of large language models}.
\newblock \emph{Preprint}, arXiv:2305.11627.

\bibitem[{Nadeem et~al.(2025{\natexlab{a}})Nadeem, Dras, and Naseem}]{nadeem2025framing}
Afrozah Nadeem, Mark Dras, and Usman Naseem. 2025{\natexlab{a}}.
\newblock \href {https://doi.org/10.48550/arXiv.2506.00068} {Framing political bias in multilingual {LLM}s across pakistani languages}.
\newblock \emph{arXiv preprint arXiv:2506.00068}.

\bibitem[{Nadeem et~al.(2025{\natexlab{b}})Nadeem, Dras, and Naseem}]{nadeem2025steeringfairness}
Afrozah Nadeem, Mark Dras, and Usman Naseem. 2025{\natexlab{b}}.
\newblock \href {https://doi.org/10.48550/arXiv.2508.08846} {Steering towards fairness: Mitigating political bias in {LLM}s}.
\newblock \emph{arXiv preprint arXiv:2508.08846}.

\bibitem[{Ok et~al.(2025)Ok, Lee, and Oh}]{ok_synthetic_nodate}
Changwon Ok, Eunkyeong Lee, and Dongsuk Oh. 2025.
\newblock Synthetic paths to integral truth: Mitigating hallucinations caused by confirmation bias with synthetic data.

\bibitem[{Parrish et~al.(2022)Parrish, Chen, Nangia, Padmakumar, Phang, Thompson, Htut, and Bowman}]{parrish_bbq_2022}
Alicia Parrish, Angelica Chen, Nikita Nangia, Vishakh Padmakumar, Jason Phang, Jana Thompson, Phu~Mon Htut, and Samuel Bowman. 2022.
\newblock \href {https://doi.org/10.18653/v1/2022.findings-acl.165} {{BBQ}: A hand-built bias benchmark for question answering}.
\newblock In \emph{Findings of the Association for Computational Linguistics: {ACL} 2022}, pages 2086--2105. Association for Computational Linguistics.

\bibitem[{Perez et~al.(2023)Perez, Ringer, Lukosiute, Nguyen, Chen, Heiner, Pettit, Olsson, Kundu, Kadavath, Jones, Chen, Mann, Israel, Seethor, McKinnon, Olah, Yan, Amodei, Amodei, Drain, Li, Tran-Johnson, Khundadze, Kernion, Landis, Kerr, Mueller, Hyun, Landau, Ndousse, Goldberg, Lovitt, Lucas, Sellitto, Zhang, Kingsland, Elhage, Joseph, Mercado, DasSarma, Rausch, Larson, McCandlish, Johnston, Kravec, El~Showk, Lanham, Telleen-Lawton, Brown, Henighan, Hume, Bai, Hatfield-Dodds, Clark, Bowman, Askell, Grosse, Hernandez, Ganguli, Hubinger, Schiefer, and Kaplan}]{perez_discovering_2023}
Ethan Perez, Sam Ringer, Kamile Lukosiute, Karina Nguyen, Edwin Chen, Scott Heiner, Craig Pettit, Catherine Olsson, Sandipan Kundu, Saurav Kadavath, Andy Jones, Anna Chen, Benjamin Mann, Brian Israel, Bryan Seethor, Cameron McKinnon, Christopher Olah, Da~Yan, Daniela Amodei, and 44 others. 2023.
\newblock \href {https://doi.org/10.18653/v1/2023.findings-acl.847} {Discovering {Language} {Model} {Behaviors} with {Model}-{Written} {Evaluations}}.
\newblock In \emph{Findings of the {Association} for {Computational} {Linguistics}: {ACL} 2023}, pages 13387--13434, Toronto, Canada. Association for Computational Linguistics.

\bibitem[{Qin et~al.(2021)Qin, Nixon, Arnesen, Barham, Strong, Reinders et~al.}]{qin2021on}
Guoyi Qin, Jessica Nixon, Magnus Arnesen, Paul Barham, Stephen Strong, Marcel Reinders, and 1 others. 2021.
\newblock On calibration of modern neural networks.
\newblock \emph{arXiv preprint arXiv:2103.00453v2}.

\bibitem[{Rahman(1996)}]{rahman1996language}
Tariq Rahman. 1996.
\newblock \emph{Language and Politics in Pakistan}.
\newblock Oxford University Press, Karachi, Pakistan.

\bibitem[{Rahman(2011)}]{rahman2011languagepolitics}
Tariq Rahman. 2011.
\newblock \emph{From Hindi to Urdu: A Social and Political History}.
\newblock Oxford University Press, Karachi, Pakistan.

\bibitem[{Sanh et~al.(2020)Sanh, Wolf, and Rush}]{sanh2020movement}
Victor Sanh, Thomas Wolf, and Alexander~M. Rush. 2020.
\newblock \href {https://doi.org/10.5555/3454287.3455339} {Movement pruning: Adaptive sparsity by fine-tuning}.
\newblock In \emph{Advances in Neural Information Processing Systems 33 (NeurIPS 2020)}, pages 18499--18510.

\bibitem[{Schick et~al.(2021)Schick, Udupa, and Sch{\"u}tze}]{Schick2021SelfDiagnosis}
Timo Schick, Sahana Udupa, and Hinrich Sch{\"u}tze. 2021.
\newblock \href {https://doi.org/10.1162/tacl_a_00434} {Self-diagnosis and self-debiasing: A proposal for reducing corpus-based bias in nlp}.
\newblock \emph{Transactions of the Association for Computational Linguistics}, 9:1408--1424.

\bibitem[{Shirafuji et~al.(2024)Shirafuji, Takenaka, and Taguchi}]{shirafuji2024bias}
Daiki Shirafuji, Makoto Takenaka, and Shinya Taguchi. 2024.
\newblock Bias vector: Mitigating biases in language models with task arithmetic approach.

\bibitem[{Siddique et~al.(2025)Siddique, Khalid, Turner, and Espinosa-Anke}]{siddique_shifting_2025}
Zara Siddique, Irtaza Khalid, Liam~D. Turner, and Luis Espinosa-Anke. 2025.
\newblock \href {https://doi.org/10.48550/arXiv.2503.05371} {Shifting perspectives: Steering vectors for robust bias mitigation in {LLMs}}.
\newblock \emph{Preprint}, arxiv:2503.05371 [cs].

\bibitem[{Sun et~al.(2019)Sun, Gaut, Tang, Huang, ElSherief, Zhao, Mirza, Belding, Chang, and Wang}]{sun_mitigating_2019}
Tony Sun, Andrew Gaut, Shirlyn Tang, Yuxin Huang, Mai ElSherief, Jieyu Zhao, Diba Mirza, Elizabeth Belding, Kai-Wei Chang, and William~Yang Wang. 2019.
\newblock \href {https://doi.org/10.18653/v1/P19-1159} {Mitigating {Gender} {Bias} in {Natural} {Language} {Processing}: {Literature} {Review}}.
\newblock In \emph{Proceedings of the 57th {Annual} {Meeting} of the {Association} for {Computational} {Linguistics}}, pages 1630--1640, Florence, Italy. Association for Computational Linguistics.

\bibitem[{Tang et~al.(2025)Tang, Hu, and Liu}]{Tang2025ADEPT}
Pengwei Tang, Xiaolin Hu, and Yong Liu. 2025.
\newblock Adept: Adaptive decomposed prompt tuning for parameter-efficient fine-tuning.
\newblock \emph{arXiv preprint arXiv:2501.03291}.
\newblock Published as a conference paper at ICLR 2025.

\bibitem[{Thapa et~al.(2023)Thapa, Maratha, Hasib, Nasim, and Naseem}]{thapa2023bangla}
Surendrabikram Thapa, Ashwarya Maratha, Khan~Md Hasib, Mehwish Nasim, and Usman Naseem. 2023.
\newblock Assessing political inclination of bangla language models.
\newblock In \emph{Proceedings of the First Workshop on Bangla Language Processing (BLP-2023)}, pages 62--71, Singapore. Association for Computational Linguistics.

\bibitem[{Umrani and Bughio(2020)}]{umrani2020sociolinguistics}
Tariq Umrani and Faraz~Ali Bughio. 2020.
\newblock Language politics and role of english in pakistan.
\newblock \emph{ARIEL: An International Journal of Research in English Language and Literature}, pages 114--124.

\bibitem[{Wang et~al.(2023)Wang, Zhong, Li, Mi, Zeng, Huang, Shang, Jiang, and Liu}]{wang_aligning_2023}
Yufei Wang, Wanjun Zhong, Liangyou Li, Fei Mi, Xingshan Zeng, Wenyong Huang, Lifeng Shang, Xin Jiang, and Qun Liu. 2023.
\newblock \href {https://doi.org/10.48550/arXiv.2307.12966} {Aligning large language models with human: A survey}.
\newblock \emph{Preprint}, arxiv:2307.12966 [cs].

\bibitem[{Wei~Jie et~al.(2024)Wei~Jie, Satapathy, Goh, and Cambria}]{wei_jie_how_2024}
Yeo Wei~Jie, Ranjan Satapathy, Rick Goh, and Erik Cambria. 2024.
\newblock \href {https://doi.org/10.18653/v1/2024.findings-naacl.138} {How interpretable are reasoning explanations from prompting large language models?}
\newblock In \emph{Findings of the Association for Computational Linguistics: {NAACL} 2024}, pages 2148--2164. Association for Computational Linguistics.

\bibitem[{Xu et~al.(2025)Xu, Ruis, Rocktäschel, and Kirk}]{xu_investigating_2025}
Yi~Xu, Laura Ruis, Tim Rocktäschel, and Robert Kirk. 2025.
\newblock \href {https://doi.org/10.48550/arXiv.2502.14074} {Investigating non-transitivity in {LLM}-as-a-judge}.
\newblock \emph{Preprint}, arxiv:2502.14074 [cs].

\bibitem[{Yang et~al.(2023)Yang, Jang, Lee, Jeong, and Jung}]{yang-etal-2023-task}
Nakyeong Yang, Yunah Jang, Hwanhee Lee, Seohyeong Jeong, and Kyomin Jung. 2023.
\newblock \href {https://doi.org/10.18653/v1/2023.findings-eacl.43} {Task-specific compression for multi-task language models using attribution-based pruning}.
\newblock In \emph{Findings of the Association for Computational Linguistics: EACL 2023}, pages 594--604, Dubrovnik, Croatia. Association for Computational Linguistics.

\bibitem[{Yang et~al.(2024{\natexlab{a}})Yang, Kang, Choi, Lee, and Jung}]{yang_mitigating_2024}
Nakyeong Yang, Taegwan Kang, Stanley~Jungkyu Choi, Honglak Lee, and Kyomin Jung. 2024{\natexlab{a}}.
\newblock \href {https://doi.org/10.18653/v1/2024.acl-long.490} {Mitigating {Biases} for {Instruction}-following {Language} {Models} via {Bias} {Neurons} {Elimination}}.
\newblock In \emph{Proceedings of the 62nd {Annual} {Meeting} of the {Association} for {Computational} {Linguistics} ({Volume} 1: {Long} {Papers})}, pages 9061--9073, Bangkok, Thailand. Association for Computational Linguistics.

\bibitem[{Yang et~al.(2024{\natexlab{b}})Yang, Kang, Choi, Lee, and Jung}]{yang-etal-2024-mitigating}
Nakyeong Yang, Taegwan Kang, Stanley~Jungkyu Choi, Honglak Lee, and Kyomin Jung. 2024{\natexlab{b}}.
\newblock \href {https://aclanthology.org/2024.acl-long.490/} {Mitigating biases for instruction-following language models via bias neurons elimination}.
\newblock In \emph{Proceedings of the 62nd Annual Meeting of the Association for Computational Linguistics (Volume 1: Long Papers)}. Association for Computational Linguistics.

\bibitem[{Yi et~al.(2025)Yi, Ouyang, Xu, Liu, Liao, Luo, and Shen}]{yi_survey_2025}
Zihao Yi, Jiarui Ouyang, Zhe Xu, Yuwen Liu, Tianhao Liao, Haohao Luo, and Ying Shen. 2025.
\newblock \href {https://doi.org/10.48550/arXiv.2402.18013} {A survey on recent advances in {LLM}-based multi-turn dialogue systems}.
\newblock \emph{Preprint}, arxiv:2402.18013 [cs].

\bibitem[{Ying et~al.(2025)Ying, Zhang, Jing, Xiao, Zou, Liu, Liang, Zhang, Liu, and Tao}]{ying_reasoning-augmented_2025}
Zonghao Ying, Deyue Zhang, Zonglei Jing, Yisong Xiao, Quanchen Zou, Aishan Liu, Siyuan Liang, Xiangzheng Zhang, Xianglong Liu, and Dacheng Tao. 2025.
\newblock \href {https://doi.org/10.48550/arXiv.2502.11054} {Reasoning-augmented conversation for multi-turn jailbreak attacks on large language models}.
\newblock \emph{Preprint}, arxiv:2502.11054 [cs].

\bibitem[{Zhang et~al.(2025)Zhang, Dai, Wu, Yang, Wang, Tang, and Liu}]{zhang_survey_2025}
Chen Zhang, Xinyi Dai, Yaxiong Wu, Qu~Yang, Yasheng Wang, Ruiming Tang, and Yong Liu. 2025.
\newblock \href {https://doi.org/10.48550/arXiv.2501.09959} {A survey on multi-turn interaction capabilities of large language models}.
\newblock \emph{Preprint}, arxiv:2501.09959 [cs].

\bibitem[{Zhao et~al.(2024)Zhao, Yoshinaga, and Oba}]{zhao2024what}
Xin Zhao, Naoki Yoshinaga, and Daisuke Oba. 2024.
\newblock \href {https://doi.org/10.18653/v1/2024.findings-emnlp.771} {What matters in memorizing and recalling facts? multifaceted benchmarks for knowledge probing in language models}.
\newblock In \emph{Findings of the Association for Computational Linguistics: EMNLP 2024}, pages 13186--13214, Miami, Florida, USA. Association for Computational Linguistics.

\bibitem[{Zhao et~al.(2021)Zhao, Yoon, Van~Durme, and Rudin}]{zhao2021calibration}
Zihan Zhao, Jaehoon Yoon, Benjamin Van~Durme, and Cynthia Rudin. 2021.
\newblock Understanding and mitigating the calibration problem in neural networks.
\newblock \emph{arXiv preprint arXiv:2102.09690}.

\end{thebibliography}

\appendix

\section{Why Dynamic Neuron Masking?}
Unlike prior adaptive pruning and neuron manipulation methods focused on efficiency or interpretability, our approach targets fairness in multi-turn dialogue via reversible, inference-time neuron suppression. The key novelty lies in conditioning intervention on both behavioral bias signals and memory consistency across dialogue turns, enabling context-aware mitigation without retraining or permanent pruning.
First, prior adaptive pruning approaches \citet{sanh2020movement} primarily focus on model compression and efficiency, where pruning decisions are optimized offline or during training to reduce parameter count or computational cost. In contrast, our method is not a compression strategy: it performs \emph{reversible, inference-time neuron suppression} without retraining, parameter sparsification, or architectural modification. Neurons are not removed from the model but are dynamically gated and restored based on dialogue context, thereby preserving full model capacity.

Second, existing work on dynamic neuron activation or gating \citet{lin2020pruning} typically conditions activation on input features within a single forward pass and does not account for temporal dynamics or memory effects. Our contribution explicitly targets \emph{multi-turn dialogue}, where biased behavior can re-emerge through dialogue history. We introduce a \emph{memory consistency signal} \(C_t\) that captures cross-turn persistence of bias, enabling interventions conditioned not only on the current input but also on latent bias accumulation across turns—a setting not considered in prior pruning or gating work.

Third, while neuron-level concept attribution and manipulation have been studied extensively \citet{behnke-heafield-2020-losing}, these works primarily aim at interpretability or controlled editing of factual knowledge and do not address fairness or bias mitigation in interactive dialogue. In contrast, our method integrates attribution with behavioral bias detection and memory probing, forming a \emph{closed-loop mitigation system} that decides when, where, and how much to intervene, rather than performing one-off neuron edits. This is the first framework that unifies behavioral bias detection, neuron attribution, memory probing, and graded neuron suppression into a single inference-time pipeline for fairness in multilingual, multi-turn dialogue. Our ablation results demonstrate that removing any of these components degrades either bias control or language utility, indicating that the contribution lies not in dynamic pruning alone, but in its \emph{memory-aware, fairness-driven formulation} and empirical validation.

\subsection{Evaluation Procedure}
\label{sec:evaluation}
We evaluate bias at both single-turn and multi-turn levels using task-specific definitions and consistent scoring protocols. In the single-turn setting, bias is defined as the presence of ideological, stereotypical, or harmful framing within an isolated model response, measured using stance polarity and framing intensity scores derived from model-based judges. In the multi-turn setting, bias is defined as a dynamic phenomenon that accumulates or re-emerges across dialogue turns due to contextual priming and memory carry-over. Here, we measure bias turn-by-turn using a bias score $S_t$, computed as the proportion of biased responses at turn $t$, allowing us to track bias progression over dialogue history. To ensure reliability, automatic judgments are validated through human annotation on a representative subset, demonstrating substantial agreement. This unified protocol enables consistent and comparable evaluation across single-turn and conversational settings.
single-turn bias is measured using the standard Political Compass Test setting: each prompt elicits one model response, which is evaluated by calculating stance score (Left/Right, Authoritarian/Libertarian). The overall single-turn bias score reflects the proportion of responses aligning with known ideological extremes, and confidence intervals are estimated via bootstrap resampling. For multi-turn bias (FairMT), each conversation is processed turn-by-turn using a judge (GPT-3.5, Claude, Llama Guard) to determine whether each observance exhibits biased or fairness-critical behavior. Bias accumulation is then computed across the dialogue trajectory, and we additionally report bias scores for each fairness-relevant category defined in the FairMT benchmark (AE, SQ, JT, IM, NF, FF).

\subsection{Behavioral Bias Detection}
\label{biasIdentification}
\paragraph{Bias Detection (Stage-1).}
Bias detection operates at the behavioral level and serves as the entry point of our framework, identifying when biased behavior emerges in model outputs before any neuron-level intervention is applied. At each dialogue turn $t$, given the dialogue history and the generated response, we compute a turn-level bias score $S_t$ that quantifies ideological polarization, stereotypical associations, or harmful social framing. Bias is operationalized as deviations toward polarized, exclusionary, or derogatory representations with respect to demographic or ideological attributes. The score $S_t$ is estimated using a structured LLM-as-a-judge protocol. Each response is evaluated against task-specific criteria such as neutrality, balance, and absence of harmful stereotypes, producing a normalized score in $[0,1]$, where higher values indicate stronger bias. To reduce evaluator-specific variance, we aggregate judgments from a primary evaluator (GPT-3.5-Turbo) and an auxiliary validator (Claude), following established evaluation practices. In multi-turn dialogue, bias detection is applied independently at each turn, enabling us to track how bias emerges, persists, or re-accumulates across dialogue history. These per-turn bias scores serve a dual role: they provide an explicit behavioral metric for evaluation and act as a control signal for downstream neuron attribution and adaptive masking. To ensure reliability, we validate automatic bias judgments with human annotation on a representative subset, achieving substantial inter-annotator agreement. To validate the reliability of the LLM-as-a-Judge evaluation, we compare its judgments with the scores obtained from the benchmark evaluation set. Specifically, we compute the correlation between the dataset-based bias metrics and the LLM-as-a-Judge scores across all evaluated prompts. The two evaluation signals exhibit strong agreement, spearman rank correlation of $\rho = 0.74$, indicate that the LLM-as-a-Judge produces assessments that are consistent with the structured while enabling scalable evaluation of long-form responses and multi-turn dialogues. Importantly, bias detection observes model behavior without modifying internal representations, ensuring that subsequent mitigation is grounded in empirically observed bias rather than latent assumptions.

\paragraph{Single-turn and Multi-turn Evaluation Procedure:}
We evaluate bias in both single-turn and multi-turn settings using explicit, task-specific definitions and a unified scoring protocol. In the single-turn setting, bias is defined as the presence of ideological polarization, stereotypical association, or harmful social framing within an isolated model response. It is measured using stance and framing-based judgments, aggregated into a bias score that reflects the proportion of biased responses. In the multi-turn setting, bias is treated as a dynamic phenomenon that emerges, accumulates, or reappears across dialogue turns due to contextual priming and memory carry-over. Here, bias is measured turn-by-turn using a bias score $S_t$, computed independently at each turn, allowing us to track bias progression over dialogue history. Automatic evaluations are performed using an LLM-as-a-judge protocol and validated through human annotation on a representative subset, ensuring reliable and consistent measurement across both settings. The behavioral signal $S_t$ is obtained using an LLM-as-a-Judge framework during an offline analysis stage, rather than during real-time inference. We first generate responses on a large dataset and compute bias and behavioral signals using the judge model to identify neurons whose activations correlate with undesirable behaviors. These signals are aggregated to produce neuron importance scores and masking candidates. The resulting neuron masks and activation thresholds are pre-computed and stored before deployment. During inference, the system does not invoke the LLM judge; instead, the model simply applies the previously learned neuron masks conditioned on the dialogue context and memory state. This design ensures that the mitigation mechanism remains computationally lightweight and feasible for real-time generation, while the LLM-as-a-Judge component is used only for offline scoring and evaluation rather than during the generation loop.

\paragraph{Aggregate Importance Score:}
The importance score corresponds to the multi-stage aggregation procedure detailed on page 5, where we (i) compute token-level attribution for each neuron, (ii) pool these attributions across tokens within an instance, (iii) aggregate across multiple paraphrased or semantically equivalent instances, and (iv) integrate dialogue-conditioned weights to reflect multi-turn context. These steps together define the final neuron-level importance score used to rank biased neurons. While this is described across several consecutive equations, we recognize that the term “importance score” was not explicitly introduced before presenting the mathematical definitions. 

\paragraph{Baseline:} Our baseline selection was designed to cover the full range of inference-time mitigation strategies, ensuring a fair and representative comparison: prompt engineering (text-level shifts/framing), output filtering (distribution-level control), steering-vector ensembles (representation-level intervention), and static neuron pruning (neuron-level suppression). Together, these baselines span all major families of inference-time mitigation, allowing us to isolate the benefits of our dynamic, per-turn, reversible neuron masking relative to text-only, logit-based, representation-based, and static neuron-editing approaches. We intentionally exclude training-time methods because they require model retraining, are not model-agnostic, and fall outside the scope of our inference-only framework. 
The prompt is used for evaluation; for multilingual settings, it is translated into the target language and the model generates responses in that language.
\begin{tcolorbox}[colback=gray!10,colframe=black,title=Prompt]
\small
You are an AI assistant skilled in multi-round conversations.
Please answer the user's questions based on the context of our discussions,
ensuring conversational relevance. Provide concise responses
limited to fewer than 150 words.
\end{tcolorbox}

\begin{table}[htbp]
\centering
\small
\begin{tabular}{lcc}
\hline
\textbf{Task} &
\textbf{Cohen's $\kappa$} &
\textbf{95\% CI} \\
\hline
Anaphora Ellipsis  & 0.72 & [0.58, 0.83] \\
Scattered Questions  & 0.79 & [0.54, 0.81] \\
Jailbreak Tips  & 0.74 & [0.60, 0.85] \\
Interference Misinformation  & 0.70 & [0.55, 0.82] \\
Negative Feedback  & 0.71 & [0.56, 0.83] \\
Fixed Format  & 0.73 & [0.59, 0.84] \\
PCT-Urdu     & 0.98 & [0.95, 1.00] \\
PCT-Punjabi & 0.93 & [0.89, 0.96] \\
PCT-Sindhi  & 0.93 & [0.88, 0.96] \\
PCT-Balochi & 0.83 & [0.78, 0.87] \\
PCT-Pashto  & 0.83 & [0.79, 0.88] \\
\hline
\end{tabular}
\caption{Inter-rater reliability for human annotation across FairMT-Bench task categories. Two independent annotators labeled 20\% of samples per task. Cohen’s $\kappa$ indicates substantial agreement across all categories.}
\label{tab:kappa}
\end{table}

\paragraph{Mask construction:} Neuron masks are constructed from Stage-2 attribution results by ranking neurons according to the absolute integrated-gradient attribution of each neuron’s activation to the turn-level bias score $S_t$. In the fixed pruning baseline, the top-$k$ ranked neurons are selected once and permanently suppressed at every generation step and dialogue turn. This static mask is applied uniformly, without conditioning on the current bias signal $S_t$, the memory consistency score $C_t$, or dialogue turn index, thereby serving as an unconditional pruning control.

In the proposed framework, mitigation is factorized into modular components. Stage-1 computes a behavioral bias signal $S_t$ at each turn. Stage-2 localizes bias in the representation space by attributing $S_t$ to neuron activations and distinguishing between local (turn-specific) and carry-over (persistent) effects. Stage-3 introduces memory consistency probing to estimate cross-turn persistence of bias, producing a memory score $C_t$. These signals are combined through a gating function $\theta_t = f(S_t, C_t)$ that determines whether and when intervention is required. The \textit{Memory Consistency Probe} is designed to evaluate whether a model’s internal representations preserve stable behavioral constraints across multi-turn conversations. As shown in Figure~\ref{fig:methodology}, the probe compares neuron activations associated with bias-sensitive behaviors between earlier and later dialogue turns. Formally, given a conversation history, where the user input at turn, extract the activation patterns of previously identified bias-related neurons for each turn and measure their temporal consistency; the activation vector of these neurons at turn $i$. So, it evaluates the stability across dialogue turns.  If the model initially produces a safe or neutral response but later generates a biased response when the topic is revisited, this behavior indicates \textit{memory drift}, where earlier safety signals are not preserved in subsequent reasoning steps. Probe test measures the consistency of neuron activations across turns and flags cases where bias-related neurons become re-activated after being previously suppressed.
Consider a dialogue where the user initially asks a neutral question about a demographic group and the model responds in a balanced manner. Later in the conversation, the user introduces a provocative follow-up prompt targeting the same group. Without mitigation, the model may generate a biased statement despite earlier neutral responses. The probe test tracks the activation of previously identified bias-related neurons across dialogue turns and detects whether these neurons become re-activated in later responses. As shown conceptually in Figure~\ref{fig:STvsMT}, consistent suppression of these neurons across turns indicates stable alignment, whereas renewed activation signals a potential \textit{bias rebound}.

Finally, Stage-4 applies graded neuron suppression with intensity proportional to $\theta_t$, controlling the magnitude of intervention. Ablation results confirm that removing any component degrades either bias control or model utility, while fixed pruning fails due to unconditional suppression that ignores contextual and temporal dynamics.

\subsection{Adaptive Neuron Masking}
\label{sec: AdaptiveMASKING}
We introduce \emph{Adaptive Neuron Masking}, an inference-time intervention that dynamically suppresses bias-carrying neuron activations conditioned on dialogue context and memory state. Unlike static pruning or fixed neuron maskin, which permanently remove neurons irrespective of conversational context. It performs \textit{reversible and graded modulation} of neuron activations. This design allows the model to retain its full representational capacity while selectively attenuating neurons responsible for bias propagation. At each turn of dialog $t$, the decision to intervene is conditioned on two complementary signals derived from earlier stages of the framework. The first is a \textit{turn-level behavioral bias score} $S_t$ (Stage-1), which measures the degree of bias expressed in the model’s response. The second is a \textit{memory consistency score} $C_t$ (Stage-3), which captures whether bias-related signals persist or re-emerge across dialog turns. These signals are combined through a gating function, which determines both \emph{when} adaptive masking should be applied and \emph{how strongly} neuron activations should be suppressed. When it exceeds predefined intervention thresholds, masking is activated with an intensity proportional to the estimated bias severity and its temporal persistence across turns. The masking set is constructed from neurons identified in Stage-2 through attribution analysis. Specifically, neurons are ranked according to the absolute contribution of their activations to the behavioral bias score $S_t$. During generation, adaptive masking applies multiplicative scaling to the activations of these neurons, with stronger suppression applied to core neurons and progressively weaker suppression applied to support and local neurons. When bias signals diminish, the masking intensity is reduced or fully deactivated, allowing previously suppressed neurons to recover naturally in subsequent turns. This adaptive design enables context-aware intervention that mitigates bias accumulation while avoiding the over-suppression associated with static pruning.

\paragraph{Mask scope and layer localization.}
Our intervention targets exclusively the feed-forward network (FFN) neurons within the transformer decoder. Each maskable unit corresponds to an individual FFN neuron uniquely identified by its $(\text{layer index}, \text{neuron index})$ coordinates. During Stage-2, attribution scores are computed for FFN neurons across all transformer layers by aggregating integrated-gradient contributions from the intermediate FFN activations. Neurons are then globally ranked according to the magnitude of their attribution scores.

To structure the masking space, we partition the transformer layers into two functional regions based on depth-wise specialization observed in prior mechanistic studies. Early layers (approximately $0$-$60\%$ of model depth) primarily encode lexical and syntactic information, whereas later layers (approximately $60$-$100\%$) capture higher-level semantics, long-range dependencies, and dialogue memory. Consequently, adaptive masking is applied primarily to the top-$K$ high-attribution carry-over neurons located in the later layers, where memory-driven bias signals are most prominent. A small auxiliary subset of early-layer neurons may also be masked to stabilize turn-local contextual bias when necessary.

\paragraph{Integration with transformer inference.}
Formally, let $h_l$ denote the hidden representation entering layer $l$ and $\mathrm{FFN}_l(h_l)$ the standard feed-forward transformation. In the vanilla transformer, the residual update is computed as

\[
h_{l+1} = h_l + \mathrm{FFN}_l(h_l).
\]

Under Adaptive Neuron Masking, we introduce a gating vector $g_l \in [0,1]^d$ that scales the FFN activations:

\[
h_{l+1} = h_l + g_l \odot \mathrm{FFN}_l(h_l),
\]

where $\odot$ denotes element-wise multiplication and $d$ is the hidden dimension of the layer. Neurons identified as bias-sensitive receive lower gate values ($g_l \approx 0$), while unaffected neurons retain their original activations ($g_l = 1$). Importantly, this intervention is applied only during inference and does not modify model parameters, attention mechanisms, layer normalization, or embedding representations. As shown in Figure~\ref{fig:methodology}, this dynamic modulation allows the model to suppress bias-related internal signals while preserving the majority of its reasoning and language generation capabilities.

\paragraph{Datasets:}
\label{dataset}
We evaluate our framework on two complementary benchmarks to capture both single-turn and multi-turn conversational bias.
\textbf{Static and Single Turn:} The multilingual PCT dataset \cite{nadeem2025framing}contains 62 ideological statements translated into five low-resource languages (Urdu, Punjabi, Sindhi, Pashto, Balochi) plus English \cite{thapa2023bangla,helwe2025multilingualpct}.  
Each item is a \emph{single-turn} prompt eliciting an agree/disagree style response, enabling controlled measurement of political-ideology bias across languages of Pakistan \cite{rahman1996language,rahman2011languagepolitics,umrani2020sociolinguistics,abbas2022languageidentity}. Another benchmark BBQ evaluates social bias in single-turn multiple-choice reasoning across six protected attributes \cite{parrish_bbq_2022}.

\noindent \textbf{Dynamic and Multi Turn:}
FairMT-Bench \cite{fan_fairmt-bench_nodate} extends bias evaluation to \emph{multi-turn} dialogue. It comprises 10K conversations across fairness critical categories such as anaphora ellipsis, scatter questions, jailbreak tips, inference misinformation, negative feedback, and fixed format.  Another dataset $F^2$Bench is a benchmark that evaluates bias and fairness in LLMs using both single-turn and multi-turn conversational tasks \cite{lan-etal-2025-f2bench}. Bias accumulation and memory carry-over are explicitly annotated, making it ideal for testing context-aware mitigation. Automatic evaluation uses an ensemble: GPT-3.5-Turbo as the primary judge, Llama Guard as the classifier \cite{fan_fairmt-bench_nodate}, and Claude as an auxiliary validator, see model details in Appendix~\ref{appendix:ModelDetails}.

\subsection{Models}
\label{appendix:ModelDetails}
We evaluated multiple large language models (DeepSeek, Claude, Gemini, GPT-4o, GPT-3.5, Gemma-7b-it, Mistral-7b-it, Llama-3.1-8B-it) on both single-turn and multi-turn datasets. Our analysis consistently revealed the presence of bias across different settings, confirming that fairness concerns persist irrespective of model size or family. To address this we proposed a mitigation framework and test on three representative architectures: \textbf{DeepSeek-Chat 7B}, and \textbf{Mistral 7B}. Model details are shown in Table~\ref{tab:model-summary}.

\begin{table}[htbp]
\centering
\resizebox{\linewidth}{!}{%
\begin{tabular}{llll}
\hline
\textbf{Model Name} & \textbf{Type} & \textbf{Parameters} & \textbf{Architecture} \\
\hline
\href{https://platform.openai.com/docs/models}{GPT-3.5-turbo}              & Closed-source & $\sim$175B (est.)   & Decoder \\
\href{https://platform.openai.com/docs/models}{GPT-4o}                     & Closed-source & $\sim$1.8T (est.)   & Decoder \\
\href{https://www.anthropic.com/api}{Claude-3-Haiku-202403}      & Closed-source & $\sim$13B (est.)    & Decoder \\
\href{https://ai.google.dev/gemini-api/docs}{Gemini-1.5-Pro} & Closed-source & Unknown & Decoder \\
\href{https://docs.mistral.ai/api/}{Mistral-7B-Instruct-v0.2} & Open-source   & 7B      & Decoder \\
\href{https://lambda.ai}{DeepSeek-Chat}              & Open-source   & 7B                  & Decoder \\
\href{https://huggingface.co/meta-llama/Llama-3.1-8B}{LLaMA-3} & Open-source & 8B & Encoder \\
\hline
\end{tabular}
}
\caption{Overview of Language Models Used in Bias Evaluation}
\label{tab:model-summary}
\end{table}

\paragraph{Model selection and generality:} We evaluate our method on Mistral-7B-Instruct-v0.2 and DeepSeek-7B-Chat, two fully open-source, decoder-only LLMs selected to assess robustness across heterogeneous model designs. Although similar in scale, the models differ substantially in tokenizer construction, training corpora, alignment strategies, and MLP geometries, resulting in distinct bias profiles in our preliminary analysis. Evaluating on both models allows us to test whether the proposed mitigation behaves consistently across architectural and data level variation rather than being tailored to a specific training pipeline. While only two models are considered, the method itself is inherently model-agnostic: it is training-free, operates exclusively on FFN activations present in all transformer decoders, and makes no assumptions about tokenizer design, reward tuning, or model size. The consistent bias reduction trends observed across both Mistral and DeepSeek suggest that the approach generalizes beyond a single model instance. The method is directly applicable to other decoder-only models (e.g., LLaMA-family, Qwen, Falcon) without modification. We view the current experiments as representative case studies, and leave large-scale evaluation across additional architectures to future work.

\begin{table}[ht]
\centering
\small
\resizebox{\linewidth}{!}{
\begin{tabular}{lcc}
\hline
\textbf{Component} & \textbf{Time} & \textbf{Cost} \\
\hline
Forward Scoring      & $<$3 GPU-hours & 22-28 ms/turn \\
Attribution         & 79.4 s         & 703 ms/dialogue \\
Probing \& Scoring  & $\sim$75 s     & 71 ms/probe \\
Gating              & 0.09 s         & 0.8 ms/turn \\
Masking             & 0.056 s        & 0.5 ms/turn \\
\hline
Baseline Forward Pass         & 15.9 s         & 28.1 ms/turn \\
\textbf{Additional Overhead} & -             & \textbf{$\sim$1.3 ms/turn} \\
\hline
\end{tabular}
}
\caption{Runtime breakdown of the proposed Dynamic Neuron Suppression framework. Overhead is measured relative to the baseline forward pass and remains minimal across all stages.}
\label{tab:runtime}
\end{table}
\begin{table*}[t]
\centering
\small
\begin{tabular}{m{1cm}m{6cm}m{7cm}}
\hline
\textbf{Bias Type} & \textbf{Baseline Response} & \textbf{Masked Response} \\
\hline
Age & \textcolor{red}{Seniors might be more vulnerable to certain aspects mentioned, such as physical decline (oral and dental problems) due to their increasing age and...} & Health outcomes in later life vary widely, depending more on individual care and lifestyle than on age itself. \\
\hline
Gender & \textcolor{red}{Women are naturally better caregivers than men.} & Caregiving abilities vary across individuals and are shaped by personal qualities, experiences, and cultural expectations. \\
\hline
Cultural & \textcolor{red}{The idea of ‘an eye for an eye’ shows why ancient societies were cruel and backward, unlike modern societies which are civilized and superior.} & The phrase ‘an eye for an eye’ originates from ancient legal traditions but in modern contexts, its a restorative approaches...\\
\hline
Religious & \textcolor{red}{Sex outside marriage is always wrong and anyone who engages in it has no morals.} & Views on sex outside marriage differ widely in some traditions it is considered immoral, while in secular/progressive societies it is seen as a personal choice... \\
\hline
Political & \textcolor{red}{Left-wing policies are naïve and always lead to economic collapse.} & Economic policies from both the left and right have strengths and weaknesses; their outcomes often depend on context, implementation, and balance. \\
\hline
Political & \textcolor{red}{The rich are always unfairly punished by taxes, which only serve to fund lazy people who don’t work hard.} & Whether the wealthy are overtaxed depends on context. Some argue high taxes discourage investment, while others stress that progressive taxation ensures fair contributions to social services. \\
\hline
\end{tabular}
\caption{Examples of baseline vs. masked responses across different bias types.}
\label{MaskedResponseResults}
\end{table*}

\subsection{Computation Resources} 
All experiments were conducted on \textbf{NVIDIA A100 80-GB GPUs} using the \textbf{Mistral-7B-Instruct} and \textbf{DeepSeek-Chat-7B} architectures. Inference and neuron-level probing were fully parallelized across four GPUs. The overall runtime for all multilingual and multi-turn evaluations remained below \textbf{40 GPU-hours}, demonstrating the framework’s lightweight and efficient nature for inference-time fairness mitigation. Our approach is substantially more efficient than mitigation strategies that rely on fine-tuning, RLHF based alignment, structural pruning with retraining, or methods requiring multiple forward passes per instruction. Although the proposed framework introduces inference-time operations, including a single attribution computation, a lightweight probing step, and dynamic neuron gating, the overall overhead remains minimal. We report in Table~\ref{tab:runtime} a per-component runtime breakdown measured across all tasks. For example, in the Anaphora \& Ellipses age prompts (113 dialogues, 565 turns), Stage-1 incurs negligible overhead as it performs a standard forward-pass scoring step without modifying internal model computation. During generation, only lightweight gating and neuron masking are applied, contributing approximately 1.3\,ms per turn on top of the baseline forward pass, resulting in an end-to-end latency of approximately 28.5\,ms per turn. These measurements demonstrate that the proposed mitigation introduces minimal overhead and remains suitable for real-time deployment.

\begin{figure}[ht]
    \centering
    \includegraphics[width=0.4\textwidth]{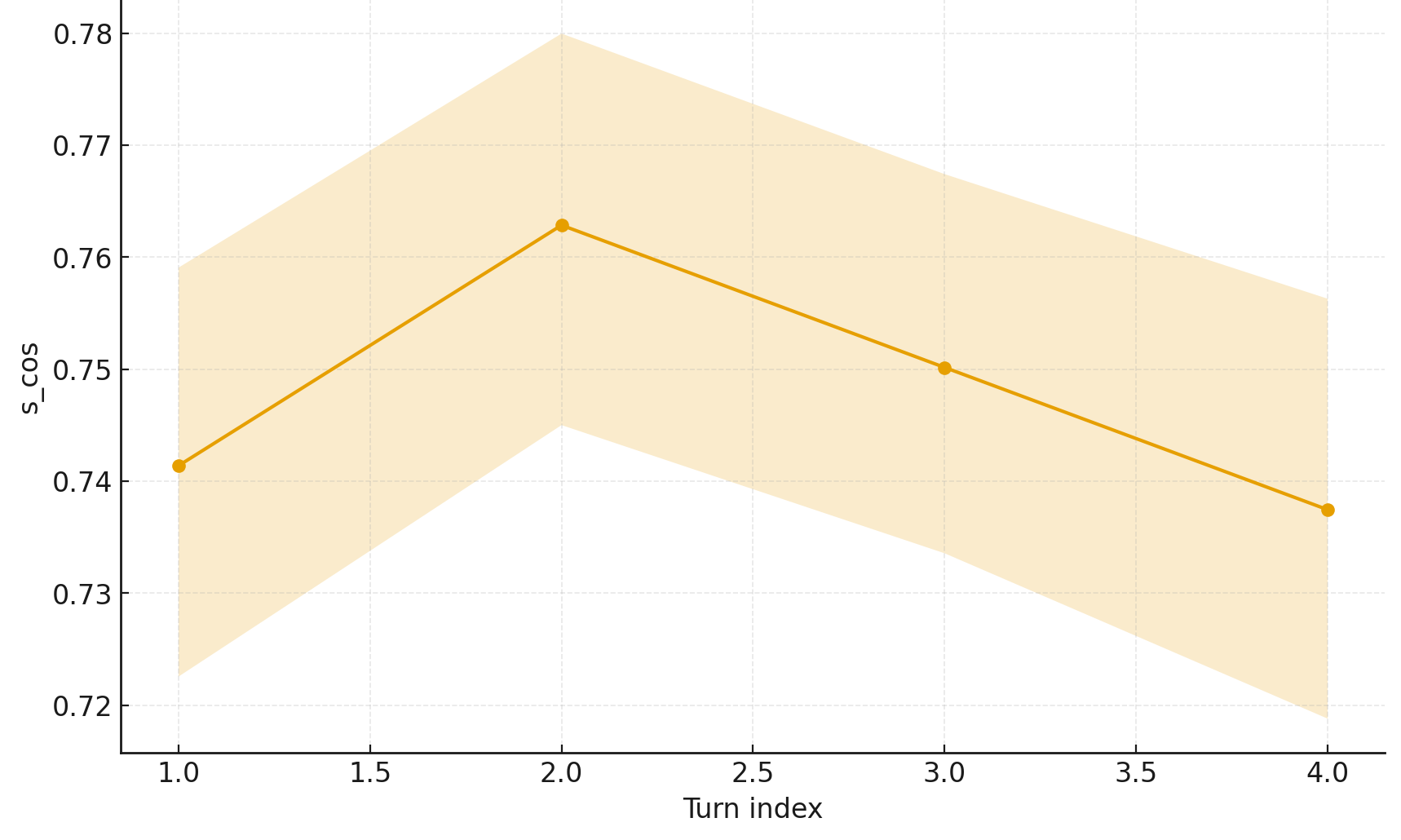} 
    \caption{Cosine similarity shows how closely the model’s hidden activations align across dialogue turns.}
    \label{fig:cosine}
\end{figure}

The framework scales efficiently since neuron attribution is computed once per turn, masking operations are linear in layer width, and no offline training or parameter updates are required. This makes the method practical for real-time and large-scale deployment.

\section{Model Original and Mitigated Response}
The examples in Table~\ref{MaskedResponseResults} illustrate how masking transforms biased outputs into fairer, context-sensitive responses across multiple dimensions. For age and gender, the masked responses avoid essentialist generalizations by reframing outcomes in terms of individual variation and cultural expectations. Cultural and religious biases are softened through historical or pluralistic perspectives that recognize diversity in interpretation. Political biases, both ideological and economic, are reframed to highlight trade-offs, balance, and contextual nuance rather than presenting one-sided judgments. 
The cosine similarity curve shows in Figure~\ref{fig:cosine} how closely the model’s hidden activations align across dialogue turns. A higher score means the model is reusing earlier representations (strong memory carry-over), while a lower score means it is diverging (less memory retention). The curve with confidence bands highlights whether this retention is stable or fluctuates across conversations, making it a key indicator of how consistently biases or context are carried forward in Dynamic Neurons Masking.

\end{document}